\documentclass[a4paper,fleqn]{cas-sc}

\usepackage[numbers]{natbib}

\usepackage{algorithm}
\usepackage{algorithmic}
\usepackage{placeins} 
\usepackage{wrapfig}  

\begin{document}
\let\WriteBookmarks\relax
\def\floatpagepagefraction{1}
\def\textpagefraction{.001}

\shorttitle{One Frame, Full Heartbeat: ECG-Free 4D Cardiac Cine MRI Synthesis}

\shortauthors{S. Wang et~al.}

\title[mode = title]{One Frame, Full Heartbeat: ECG-Free 4D Cardiac Cine MRI Synthesis via Radial-Decomposed Flow Matching}

\author[1,2]{Shiyi Wang}
\ead{swang103@mgh.harvard.edu}
\credit{Conceptualization, Methodology, Software, Validation, Writing - Original draft}

\author[2]{Ruochen Sun}
\ead{rsun6@mgh.harvard.edu}
\credit{Data Curation}

\author[3]{Xiang Li}
\ead{xli60@mgh.harvard.edu}
\credit{Formal analysis}

\author[1]{Peirong Liu}
\cormark[1]
\ead{peirong@jhu.edu}
\credit{Writing - Review \& Editing}

\author[2]{Fangxu Xing}
\cormark[1]
\ead{fxing1@mgh.harvard.edu}
\credit{Supervision}

\affiliation[1]{organization={Department of Electrical and Computer Engineering, Johns Hopkins University},
    city={Baltimore},
    state={MD},
    country={USA}}

\affiliation[2]{organization={Gordon Center for Medical Imaging, Department of Radiology, Massachusetts General Hospital and Harvard Medical School},
    city={Boston},
    state={MA},
    country={USA}}

\affiliation[3]{organization={Center for Advanced Medical Computing and Analysis, Department of Radiology, Massachusetts General Hospital and Harvard Medical School},
    city={Boston},
    state={MA},
    country={USA}}

\cortext[cor1]{Corresponding author}

\begin{abstract}
Cine cardiovascular magnetic resonance (CMR) captures the full cardiac cycle as a four-dimensional (4D) sequence, but standard acquisition depends on electrocardiogram (ECG) gating and repeated breath holds, which is difficult in uncooperative populations, resource-limited settings, and temporally corrupted datasets. Existing synthesis approaches prioritize visual realism, which alone does not establish accurate patient-specific ejection fraction (EF) or ventricular volumes.
We present PhaseFlow3D, a generative framework that synthesizes a complete 4D cine sequence from a single end-diastolic (ED) three-dimensional (3D) volume without ECG dependency. To capture the asymmetric dynamics of systole and diastole, PhaseFlow3D encodes the cardiac cycle as a piecewise linear phase anchored at the ED and end-systolic (ES) time points; at inference, this phase is instantiated from a population-level canonical template, requiring no patient-specific temporal information. Conditioned on this phase, a rectified flow model synthesizes a cardiac motion trajectory in latent space. A Radial Contraction Decomposition then decodes each latent state into a 3D displacement field, combining a physics-informed radial component for centripetal myocardial contraction with an image-conditioned residual for rotation and out-of-plane motion. Every frame is generated by directly warping the ED volume, bypassing variational autoencoder decoding.
On the combined ACDC and M\&Ms benchmark, PhaseFlow3D achieves the best EF mean absolute error, the only positive left-ventricular volume-curve $R^2$, and the best distributional quality among all compared methods, with ablations confirming the contribution of each component.
Downstream evaluations further confirm the utility of the synthesized sequences and their displacement fields in segmentation, pathology classification, label propagation, and myocardial strain analysis.
\end{abstract}

\begin{highlights}
\item ECG-free 4D cardiac cine MRI synthesis from a single ED 3D volume.
\item Piecewise linear cardiac phase with a canonical template for fully blind inference.
\item Radial Contraction Decomposition combines a physics prior with a learned residual and directly warps the real ED volume.
\item Synthetic supervision retains $99.2\%$ of real-supervision segmentation Dice.
\item Best EF accuracy, only positive volume-curve $R^2$, and best FID/FVD on ACDC and M\&Ms.
\end{highlights}


\maketitle

\section{Introduction}
\label{sec:introduction}

Cine cardiovascular magnetic resonance (CMR) imaging is the clinical gold standard
for quantifying ventricular anatomy and
function~\cite{bellenger2000interchangeable, grothues2002reproducibility},
enabling precise physiological measurements throughout the complete cardiac
cycle such as ejection fraction (EF), end-diastolic (ED) and end-systolic (ES)
volumes, myocardial strains, etc.
In standard clinical practice, a full four-dimensional (4D) CMR acquisition often requires electrocardiogram (ECG) gating and multi-breath-hold pulse sequences, a
protocol unavailable in retrospective cohort studies, prohibitively costly
in resource-limited settings, challenging for uncooperative population, and inapplicable when only isolated three-dimensional (3D) volumes
survive archiving.
Therefore, the problem of synthesizing a complete, physiologically coherent 4D cine CMR
sequence from a single 3D volume is both clinically
compelling and technically formidable: it demands recovering the full
spatiotemporal trajectory of 3D myocardial deformation from purely spatial information,
without any explicit temporal signal.
Beyond restoring the missing temporal dimension, synthesizing the full sequence
from a single volume would enable quantitative motion analysis from the recovered
displacement field, support data augmentation for data-scarce downstream tasks
such as segmentation and disease classification, and, since every generated
sequence carries a known deformation field, provide ground-truth benchmarks for
validating deformable registration algorithms.

Prior work provides several approaches to cardiac motion modeling and synthesis.
\emph{Pairwise registration} methods such as VoxelMorph~\cite{voxelmorph}
estimate deformations between observed images. Krebs et al.~\cite{krebs2019}
additionally learn a temporally regularized probabilistic motion model that
supports motion simulation from an ED image. Temporal and regional motion
modeling is also central to cardiac tracking methods such as
GPTrack~\cite{gptrack}; separate studies of learning-based registration identify
limitations in cross-domain generalization~\cite{regmagic}.
\emph{Deformation-based} generators such as DragNet~\cite{dragnet} predict
displacement fields autoregressively from a single frame. This removes the
target-frame requirement, although recurrent predictions may accumulate error
over long sequences.
\emph{ECG-conditioned} methods~\cite{echopulse, ecgflowcmr} use ECG as a
temporal conditioning signal, requiring an ECG input at inference that may be
unavailable in retrospective or resource-limited settings.
\emph{Video generators} based on generative adversarial networks
(GANs)~\cite{goodfellow2014gan} and diffusion
models~\cite{ho2020ddpm, song2021score} offer further synthesis strategies,
as surveyed in~\cite{biomedvid}. Cardiac applications include physiological
attribute editing~\cite{gancmri}, multimodal structural and motion
conditioning~\cite{heartbeat}, and ED semantic-map conditioning~\cite{echodiff_icassp}.
Conditional image leakage has also been identified as a source of insufficient
motion in image-to-video diffusion~\cite{imgleak}, underscoring the need to
evaluate dynamics alongside image appearance.
Visual realism alone does not establish the accuracy of patient-specific EF
or left-ventricular (LV) volumes. We therefore target single-volume 4D cine
synthesis with explicit supervision of EF and the LV volume trajectory.

Our temporal conditioning strategy uses cardiac phase to explicitly align ED
and ES across subjects. This compact representation accommodates the different
durations of systole and diastole~\cite{fukuta2008} without an ECG input.
Exploiting this phase signal to synthesize a complete 3D volumetric cine
sequence from a single ED volume, however, is not straightforward; it raises
three challenges.
First, the motion to be synthesized becomes a dense 3D displacement field,
whose coordinated volumetric motion is not explicitly constrained by
independent per-slice warps.
Second, the computational and memory cost of jointly modeling
$T \times D \times H \times W$ voxels precludes naive 4D architectures.
Third, physiological supervision must act on volumetric quantities (EDV, ESV,
EF) rather than per-slice areas, and must do so without ever running a
segmentation network on synthesized images.

We present \textbf{PhaseFlow3D}, a generative framework that synthesizes a
complete 4D cine CMR sequence from a single ED 3D volume. At its core, an
ECG-free cardiac phase drives a latent flow-matching architecture. The phase is
piecewise linear with breakpoints at the known ED and ES timings, encoding each
frame as $\mathbf{p}_t{=}[\sin\phi_t,\,\cos\phi_t]$. At inference, a
population-level canonical template replaces the per-patient timing,
enabling fully blind synthesis with no ECG, gating, or pathology label.
Conditioned on this phase, a rectified flow model~\cite{liu2022rectified}
is trained on straight interpolations between the ED anchor's latent features
$\mathbf{z}_0$ and the full-sequence latent features $\mathbf{z}_1$.
At inference, we evaluate the velocity network once at the anchor and decode
its output into the motion sequence. A D$'$-fold trick then folds the compressed depth
axis into channels, so 4D reasoning runs at 3D computational cost. Each
predicted latent velocity is decoded through a Radial Contraction
Decomposition: a physics-informed radial component encodes centripetal LV
contraction through an anatomy-derived direction field, ED anchoring,
phase-conditioned temporal modulation, and anatomical localization,
while an image-conditioned residual decoder synthesizes the remaining
non-radial motion. The total displacement field warps the ED volume directly
into every output frame. Output intensities are interpolated from the real
scan, bypassing variational autoencoder (VAE)~\cite{kingma2013vae} decoding.

Our main contributions are:
\begin{itemize}
  \item \textbf{An ECG-free cardiac phase for single-volume synthesis.}
        A piecewise linear cardiac phase anchored only at ED and ES replaces
        the ECG signal, and a population-level canonical template supplies
        these timings at inference, enabling fully blind synthesis from a
        single ED volume.
  \item \textbf{Radial Contraction Decomposition.}
        Cardiac motion is decomposed into a physics-informed radial component
        and a data-driven residual, and every frame is synthesized by warping
        the real ED volume, bypassing VAE decoding. To our knowledge, this is
        the first use of a radial decomposition as a structural prior in a
        generative synthesis framework.
  \item \textbf{Physiological supervision without segmenting synthesized images.}
        EF and volume-curve losses are computed by warping the ground-truth ED
        mask through the predicted displacement field, so segmentation runs
        only on real images and clinically relevant metrics supervise training
        without domain shift.
  \item \textbf{Validated downstream utility.}
        Five downstream experiments show utility beyond visualization:
        synthetic supervision can largely substitute for real supervision in
        segmentation and pathology classification, and the recovered
        displacement fields support label propagation and blind myocardial
        strain analysis.
\end{itemize}

The remainder of this paper is organized as follows.
Section~\ref{sec:related} reviews related work on cardiac motion synthesis.
Section~\ref{sec:methods} details the proposed framework.
Section~\ref{sec:experiments} presents the experimental setup, comparisons with
state-of-the-art baselines, and ablation studies.
Section~\ref{sec:downstream} evaluates the utility of the synthesized sequences
and their displacement fields in five downstream applications.
Section~\ref{sec:discussion} discusses the physiological findings and the
limitations of the method, and Section~\ref{sec:conclusion} concludes the paper
and outlines directions for future work.

\section{Related Work}
\label{sec:related}

\paragraph{Cine CMR synthesis.}
Prior cine synthesis methods differ in their conditioning inputs and motion representations.
VoxelMorph~\cite{voxelmorph} estimates pairwise deformations, whereas the
probabilistic motion model of Krebs et al.~\cite{krebs2019} also supports
sampling motion from an ED image and computing LV volume curves from warped
ED masks. Its normalized time input does not itself impose constant motion velocity.
DragNet~\cite{dragnet} predicts displacement fields autoregressively from a
single frame without a target image or ECG input; recurrent prediction may
accumulate error over long sequences.
ECG-conditioned models~\cite{echopulse, ecgflowcmr} use an ECG input to guide
video generation. GANcMRI~\cite{gancmri} explores physiological attribute
control through latent-space prompting, while echocardiographic diffusion
models use structural and motion conditions~\cite{heartbeat} or an ED
semantic map~\cite{echodiff_icassp}.
CardioDiT~\cite{cardiodit} jointly models space and time in a latent diffusion
transformer and evaluates volumetric cardiac function in unconditional 4D CMR
generation; its reported experiments do not address conditioning on a single
observed volume. PhaseFlow3D targets this conditional setting with explicit
EF and volume-curve supervision and parallel, phase-conditioned deformation
prediction. Section~\ref{sec:setup} describes our baseline adaptations.

\paragraph{Flow matching and video prediction.}
Flow matching~\cite{lipman2022flow} and rectified flow~\cite{liu2022rectified}
learn continuous transport fields, with rectification providing a route toward
efficient sampling. Straight training interpolations alone do not guarantee
an exact single-step inference trajectory.
Separately, SF-V~\cite{sfv} obtains single-forward-pass video generation by
adversarially fine-tuning a pretrained video diffusion model.
EchoLVFM~\cite{echolv} uses MeanFlow-based latent video generation conditioned
on observed echocardiographic frames and a target EF. Its 2D ultrasound setting
and EF-conditioned input differ from our single-volume CMR setting, which
requires no target EF at inference.
General-purpose video prediction models such as SimVP~\cite{simvp} and latent
flow diffusion models (LFDM)~\cite{lfdm} deliver strong pixel-level fidelity on
natural videos. Their original formulations do not include cardiac EF or
volume-curve supervision; our experiments evaluate their physiological
accuracy when adapted to single-volume cardiac synthesis.

\paragraph{Diffeomorphic deformation in medical imaging.}
Scaling-and-squaring integration~\cite{arsigny2006} provides a
computationally efficient route to diffeomorphic transformations.
VoxelMorph~\cite{voxelmorph} established a learning-based framework for
deformable registration. Dalca et al.~\cite{dalca2018} developed a probabilistic
diffeomorphic registration model evaluated on 3D brain MRI, while earlier
work incorporated incompressibility constraints for cardiac tissue
tracking~\cite{Mcircreg}.
PhaseFlow3D uses velocity integration in its generative residual decoder.
The residual is combined additively with a radial displacement, so the
regularity of the total transformation is evaluated numerically rather than
inferred solely from the residual parameterization.

\paragraph{Physics-informed cardiac motion priors.}
LV motion includes radial contraction, longitudinal shortening, and
rotation~\cite{radialmotion}. We use a radial field defined relative to the
LV centroid as a simplified anatomical prior rather than a complete
biomechanical model.
Prior registration methods incorporate tissue incompressibility through
constrained deformation models: iLogDemons~\cite{Mcircreg}, for example,
imposes a divergence-free velocity constraint. Tissue incompressibility does
not imply a constant LV cavity volume.
Our radial branch encodes an anatomy-derived direction within a generative
decoder, while a learned residual represents rotation, shear, and
out-of-plane motion.

\section{Methodology}
\label{sec:methods}

We leverage the generative nature of latent rectified flow
matching~\cite{liu2022rectified}, in combination with a physics-informed Radial
Contraction Decomposition, to introduce PhaseFlow3D for 4D cine synthesis from a
single ED volume.
Here, the purpose of PhaseFlow3D is to predict a phase-conditioned latent
velocity and decode it into a 3D displacement field
$\boldsymbol{\varphi}_t$, which warps the ED volume $\mathbf{x}_{\mathrm{ED}}$
into each synthesized frame $\hat{\mathbf{x}}_t$.
The temporal structure comes from a piecewise linear cardiac phase with
breakpoints at ED and ES, so that no ECG or per-patient temporal signal is
required.
Note that this phase can be instantiated either from ground-truth ED and ES
landmarks or, for fully blind inference, from a population-level canonical
template (Sec.~\ref{sec:phase}).
Details of the method are described in the following, including the task
formulation, the phase conditioning, the latent encoding, the phase-conditioned
flow matching, the radial displacement decoding, and the network architecture
and implementation (Sec.~\ref{sec:architecture_impl}).

\subsection{Problem Formulation}
\label{sec:problem}

Let $\mathbf{x}_{\mathrm{ED}} \in \mathbb{R}^{1 \times D \times H \times W}$
denote the single 3D ED CMR volume, with $D$ slices and $H \times W$ in-plane
resolution.
The synthesized 4D cine sequence
$\{\hat{\mathbf{x}}_1, \ldots, \hat{\mathbf{x}}_T\}$ (frame index
$t \in \{1,\dots,T\}$, with $T$ the number of frames in the cardiac cycle, each
$\hat{\mathbf{x}}_t \in \mathbb{R}^{1 \times D \times H \times W}$) must be
both \emph{perceptually realistic} and \emph{physiologically faithful}.
This task is fundamentally underdetermined: a single ED volume does not uniquely
specify the subsequent contraction trajectory. CMR functional assessment uses
measurements across the cardiac cycle~\cite{jiang2024cmr}; in echocardiography,
video-based models likewise exploit temporal information for functional
assessment~\cite{echonet}. We supply a piecewise linear cardiac phase as a
temporal prior (Sec.~\ref{sec:phase}), without assuming that it identifies a
patient's unique true motion from the ED anatomy alone.

\subsection{Piecewise Linear Cardiac Phase}
\label{sec:phase}

Providing a temporally structured conditioning signal without ECG is the
central obstacle to single-volume cine synthesis.
We parameterize the cardiac cycle by a \emph{piecewise linear phase} with
breakpoints at the two physiologically meaningful landmarks of the cycle, ED and
ES.

\paragraph{Phase definition.}
Let $t_{\mathrm{ED}}$ and $t_{\mathrm{ES}}$ denote the frame indices of
end-diastole and end-systole, respectively.
The cardiac phase is defined as
\begin{equation}
  \phi_t =
  \begin{cases}
    \pi \cdot \dfrac{t - t_{\mathrm{ED}}}{t_{\mathrm{ES}} - t_{\mathrm{ED}}} &
      \text{if } t_{\mathrm{ED}} \leq t \leq t_{\mathrm{ES}} \\[6pt]
    \pi + \pi \cdot \dfrac{t - t_{\mathrm{ES}}}{T - t_{\mathrm{ES}}} &
      \text{if } t_{\mathrm{ES}} < t \leq T
  \end{cases}
  \label{eq:phase}
\end{equation}
so that $\phi_{t_{\mathrm{ED}}} = 0$, $\phi_{t_{\mathrm{ES}}} = \pi$, and
$\phi_T = 2\pi$.
Each frame is then encoded as
$\mathbf{p}_t = [\sin\phi_t,\;\cos\phi_t] \in \mathbb{R}^2$,
a compact, periodicity-aware representation.
During training, the ED and ES positions come from ground-truth annotations.

\paragraph{Why piecewise linear?}
\begin{wrapfigure}{r}{0.40\textwidth}
  \centering
  \includegraphics[width=0.38\textwidth]{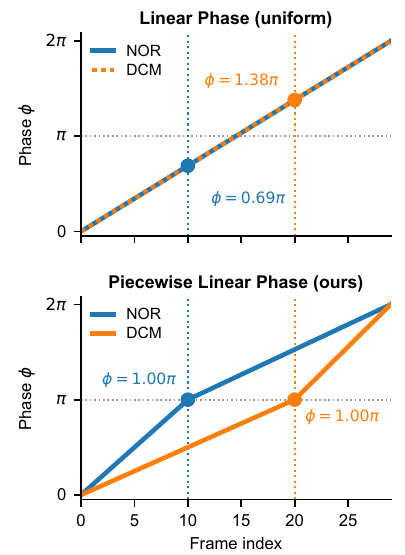}
  \caption{Uniform linear phase (top) scatters ES across $\phi$ values;
    the piecewise linear phase (bottom) aligns ES at $\phi\!=\!\pi$ by
    construction across pathological types.}
  \label{fig:toy_phase_b}
\end{wrapfigure}
Systole and diastole occupy different portions of the cardiac cycle, with
durations that vary with physiological conditions~\cite{fukuta2008}.
A uniform linear phase $\phi_t^{\text{lin}} = 2\pi t / T$ uniformly spaces
frames but does not explicitly align ES across subjects
(Figure~\ref{fig:toy_phase_b}, top).
The piecewise linear phase instead maps ES to $\phi = \pi$ by
construction, allocating a fixed phase interval to each part of the cycle
(Figure~\ref{fig:toy_phase_b}, bottom). This aligns temporal landmarks; the
learned deformation model determines the motion velocity within each interval.

\paragraph{Canonical phase for fully blind inference.}
At inference, only the ED volume is available and $t_{\mathrm{ES}}$ is unknown.
For each training subject $i$, let the ES fraction be
$r_{\mathrm{ES}}^{(i)} = t_{\mathrm{ES}}^{(i)} / T^{(i)}$, the position of ES as a
fraction of that subject's cycle length.
We pre-compute the population-average fraction
$\bar{r}_{\mathrm{ES}} = \tfrac{1}{N} \sum_{i=1}^{N} r_{\mathrm{ES}}^{(i)}$ over
the $N$ training subjects, and define the canonical phase template
\begin{equation}
  \phi_t^{\mathrm{can}} = \phi_t\bigl(t_{\mathrm{ES}} = \lfloor \bar{r}_{\mathrm{ES}}\,T \rfloor\bigr).
  \label{eq:canonical}
\end{equation}
The approximation error introduced by the canonical template is quantified in
Sec.~\ref{sec:phase_validation}.

\subsection{CineVAE3D: Per-Frame 3D Encoding}
\label{sec:vae}

A pretrained, frozen CineVAE3D encoder $E_{\mathrm{vae}}$ constructs
the latent space.
Rather than encoding an entire sequence jointly, CineVAE3D encodes each 3D
frame independently:
\begin{equation}
  \mathbf{z}_1^{(t)} = E_{\mathrm{vae}}(\mathbf{x}_t) \in \mathbb{R}^{C \times D' \times h \times w},
  \label{eq:z1}
\end{equation}
where $C$ is the number of latent channels and $D', h, w$ are the compressed
depth and in-plane sizes, so the full-sequence latent is
$\mathbf{z}_1 \in \mathbb{R}^{C \times T \times D' \times h \times w}$.
The encoder uses true 3D convolutions with per-frame 2D spatial attention at
the bottleneck.
The anchor latent is built from the ED volume repeated $T$ times with a small
additive perturbation:
\begin{equation}
  \mathbf{z}_0 = \bigl[E_{\mathrm{vae}}(\mathbf{x}_{\mathrm{ED}})\bigr]_T + \alpha\,\boldsymbol{\varepsilon},
  \quad \boldsymbol{\varepsilon} \sim \mathcal{N}(\mathbf{0},\,\mathbf{I}),
  \label{eq:z0}
\end{equation}
where $[\cdot]_T$ denotes expansion along the temporal dimension.
This division of labor is deliberate: the VAE specializes in spatial
representation, while all temporal dynamics of the transport
from $\mathbf{z}_0$ to $\mathbf{z}_1$ are delegated to the flow-matching
model.

\subsection{Phase-Conditioned Latent Flow Matching}
\label{sec:flow}

We formulate 4D cine synthesis as conditional transport in the shared latent
space using rectified flow~\cite{liu2022rectified}.

\paragraph{Rectified flow.}
Given a uniformly distributed flow time $\tau \sim \mathcal{U}(0, 1)$, the
interpolated latent feature at $\tau$ is
$\mathbf{z}_\tau = (1 - \tau)\,\mathbf{z}_0 + \tau\,\mathbf{z}_1$.
Differentiating with respect to $\tau$ gives the ground-truth feature velocity
$\mathbf{v}^* = {d\mathbf{z}_\tau}/{d\tau} = \mathbf{z}_1 - \mathbf{z}_0$,
which is constant along each sampled training interpolation.
A VelocityUNet4D $v_\theta$ is trained to predict this velocity,
conditioned on the flow time $\tau$ and the phase tokens $\mathbf{c}$:
\begin{equation}
  \hat{\mathbf{v}} = v_\theta(\mathbf{z}_\tau,\;\tau,\;\mathbf{c}),
  \qquad
  \mathbf{c} = \mathbf{e}_{\mathrm{phase}} \in \mathbb{R}^{T \times d},
  \label{eq:velocity}
\end{equation}
where $\mathbf{e}_{\mathrm{phase}}$ is our proposed per-frame phase embedding
produced by a small multilayer perceptron (MLP; the PhaseProjector) from
$\mathbf{p}_t$ and injected at every resolution level via
cross-attention~\cite{vaswani2017attention}, enabling spatially adaptive phase
conditioning.
The global condition for adaptive group normalization
(AdaGN)~\cite{adagn} is taken as the temporal mean of
$\mathbf{e}_{\mathrm{phase}}$.

\paragraph{VelocityUNet4D and the D$'$-fold trick.}
The second obstacle to single-volume cine synthesis is computational: naive 4D
convolutions over the full $(T, D', h, w)$ latent grid are prohibitively
expensive in both memory and compute. We resolve this by exploiting the fact
that the VAE has already compressed the depth axis to only $D'$ slices, which
is shallow enough to fold into the channel dimension without inflating the
network. Folding the $D'$ depth slices into the channel dimension turns the 4D
problem into a 3D one over $(T, h, w)$, with the depth slices mixed through the
channel weights. The folded tensor is
processed by a 3D U-Net backbone
and unfolded back to the 4D latent after each resolution level. All attention
blocks thus operate without modification, and full 4D reasoning is achieved at
the cost of a 3D network.
To enable cross-frame reasoning, each resolution level incorporates a temporal
attention block that exchanges information across all $T$ frames
simultaneously, allowing the model to synthesize temporally coherent motion
rather than processing each frame independently.
At inference, we evaluate the velocity network once at the ED-conditioned anchor:
$\hat{\mathbf{v}} = v_\theta(\mathbf{z}_0,\,0,\,\mathbf{c})$,
and decode its output directly into the motion sequence without iterative
ordinary differential equation (ODE) integration. This single-evaluation
scheme is an inference design choice evaluated in our experiments; the
constant target velocity on training interpolations does not imply that the
learned field is constant along its inference trajectories.

\subsection{Radial Contraction Decomposition}
\label{sec:radial}

The predicted latent velocity $\hat{\mathbf{v}}$ must be decoded into a
physiologically meaningful 3D displacement field.
Instead of an unconstrained decoder, we decompose the displacement into two
complementary components:
\begin{equation}
  \boldsymbol{\varphi}_t = \boldsymbol{\varphi}_t^{\mathrm{radial}}
  + \boldsymbol{\varphi}_t^{\mathrm{residual}},
  \label{eq:decomp}
\end{equation}
where the radial component provides a simplified prior for LV contraction and
the residual models additional motion.
Throughout, $\boldsymbol{\varphi}_t$ denotes the backward sampling map used by
the spatial transformer (Sec.~\ref{sec:warp}): the output voxel at location
$\mathbf{x}$ draws its intensity from location
$\mathbf{x} + \boldsymbol{\varphi}_t(\mathbf{x})$ in the ED volume.
Under this convention, a sampling displacement directed away from the LV
centroid retrieves intensities from larger radii, so the rendered tissue
moves toward the centroid.

\paragraph{Radial component (physics-informed prior).}
Radial contraction is one component of the complex LV motion, alongside
longitudinal shortening and rotation~\cite{radialmotion}. We approximate this
component with a unit direction field defined relative to the LV centroid.
We define the radial direction field as
\begin{equation}
  \mathbf{r}(\mathbf{x}) = \frac{\mathbf{x} - \mathbf{x}_{\mathrm{LV}}}
    {\|\mathbf{x} - \mathbf{x}_{\mathrm{LV}}\| + \varepsilon}
    \cdot \mathbf{M}_{\mathrm{dil}},
  \label{eq:radial_field}
\end{equation}
where $\mathbf{x}$ is a voxel coordinate, $\mathbf{x}_{\mathrm{LV}}$ is the LV
centroid, $\varepsilon$ is a small constant that prevents division by zero, and
$\mathbf{M}_{\mathrm{dil}}$ is a dilated binary LV mask that confines the radial
field to the cardiac region.
The radial displacement at frame $t$ is then
\begin{equation}
  \boldsymbol{\varphi}_t^{\mathrm{radial}} =
    A_{\mathrm{pat}} \cdot \mathbf{m}_{\mathrm{loc}}(\mathbf{x}) \cdot s_t \cdot \mathbf{r}(\mathbf{x}),
  \label{eq:radial}
\end{equation}
where
$A_{\mathrm{pat}} = \sigma(\mathrm{GAP}(\mathbf{f}))\cdot A_{\max}$
is a patient-specific global amplitude derived from the ED image features
$\mathbf{f}$ via global average pooling, with $\sigma$ the sigmoid function and
$A_{\max}$ an upper bound on the radial amplitude;
$\mathbf{m}_{\mathrm{loc}}(\mathbf{x}) = \sigma(\mathrm{LocalNet}(\mathbf{f}))$
is a spatially adaptive modulation map weighting the radial contribution
across myocardial regions; and
$s_t = \mathrm{peak\_norm}\bigl(\sigma(\mathrm{TemporalNet}(\mathbf{e}_{\mathrm{phase}}^t)) - s_0\bigr)$
is a temporal activation curve, where $\mathrm{peak\_norm}$ rescales it to unit
peak so that $s_0 = 0$ at ED and $s_t = 1$ at the systolic peak.
This branch incorporates four design properties:
(1) a radial direction prior: for nonnegative $s_t$, the sampling
displacement points away from the centroid, which renders centripetal
tissue motion under the backward-sampling convention;
(2) ED identity: $s_0 = 0$ enforces
$\boldsymbol{\varphi}^{\mathrm{radial}}_{t_{\mathrm{ED}}} = \mathbf{0}$;
(3) continuous temporal modulation through the phase-conditioned
sigmoid-MLP parameterization; and
(4) spatial localization: $\mathbf{M}_{\mathrm{dil}}$ confines radial
motion to the cardiac region.

\paragraph{Residual component.}
The residual is decoded from $\hat{\mathbf{v}}$ by an
ImageConditionedDeformDecoder3D, whose upsampling layers receive
skip-connected multi-scale features $\mathbf{f}^{1:3}$ of the ED volume from a
lightweight ImageEncoder3D:
\begin{equation}
  \boldsymbol{\varphi}^{\mathrm{residual}} =
    \mathrm{ScalingSquaring3D}\bigl(D_{\mathrm{deform}}(\hat{\mathbf{v}},\,\mathbf{f}^{1:3})\bigr).
  \label{eq:residual}
\end{equation}
Scaling-and-squaring integration~\cite{arsigny2006} with $K$ steps approximates
the flow of a stationary velocity field, providing a diffeomorphic
parameterization in the continuous setting. Finite-step integration and
discrete interpolation do not give an unconditional guarantee of positive
Jacobian determinants. Moreover, adding the radial displacement does not
preserve such a guarantee for the total field. We therefore assess its
folding rate empirically in Sec.~\ref{sec:diffeomorphism}.
Throughout, $\mathbf{J}_{\boldsymbol{\varphi}}$ denotes the Jacobian of the
coordinate transformation $\mathrm{Id}+\boldsymbol{\varphi}$, namely
$\mathbf{I}_3+\nabla\boldsymbol{\varphi}$, rather than the derivative of the
displacement alone.
The skip connections ground the residual in patient anatomy: the decoder
observes the actual ventricular boundaries when synthesizing non-radial
motion such as rotation, shear, and out-of-plane deformation.

\paragraph{Residual regularization.}
To discourage the residual from canceling the radial component, we penalize its
projection onto the radial direction:
\begin{equation}
  \mathcal{L}_{\mathrm{res}} =
    \Bigl|\,\bigl(\boldsymbol{\varphi}^{\mathrm{residual}} \cdot \mathbf{r}(\mathbf{x})\bigr)\Bigr|_{\mathrm{mean}},
  \label{eq:lres}
\end{equation}
encouraging the radial head to account for radial contraction.

\subsection{Warp-Based 4D Synthesis}
\label{sec:warp}

Given the total displacement $\boldsymbol{\varphi}_t$, each output frame is
synthesized by directly warping the ED volume:
\begin{equation}
  \hat{\mathbf{x}}_t = \mathrm{warp}(\mathbf{x}_{\mathrm{ED}},\,\boldsymbol{\varphi}_t),
  \label{eq:warp}
\end{equation}
where $\mathrm{warp}$ is a differentiable 3D spatial transformer with
trilinear grid sampling~\cite{jaderberg2015stn}, evaluating
$\hat{\mathbf{x}}_t(\mathbf{x}) =
\mathbf{x}_{\mathrm{ED}}\bigl(\mathbf{x} + \boldsymbol{\varphi}_t(\mathbf{x})\bigr)$
at every output voxel $\mathbf{x}$.
The map $\mathbf{x} \mapsto \mathbf{x} + \boldsymbol{\varphi}_t(\mathbf{x})$
therefore sends frame-$t$ coordinates to ED coordinates.
Forward tissue motion corresponds to the inverse of this map, a distinction
the strain analysis accounts for by inverting the deformation gradient
(Sec.~\ref{sec:strain}).
Every synthesized voxel is an interpolation of the real ED volume, and no VAE
decoder participates in the output synthesis path. This anchors image
appearance to the observed anatomy and requires temporal variation to be
expressed through a displacement field. It does not eliminate interpolation
error or guarantee anatomically correct motion; the physiological losses and
deformation regularization constrain the predicted fields. The effect of
replacing this warp-based synthesis with VAE decoding is evaluated in
Sec.~\ref{sec:ablation} (A1).

\subsection{Training Objectives}
\label{sec:loss}

PhaseFlow3D is trained end-to-end with a composite objective that combines a
core flow-matching term in the latent space, an image-space fidelity term, and a
set of physiological and anatomical constraints. Each component is described
below, and the terms are weighted into a single total loss.

\paragraph{Flow matching loss.}
The standard rectified flow matching objective, evaluated in latent space:
\begin{equation}
  \mathcal{L}_{\mathrm{flow}} = \mathbb{E}_{\tau,\,(\mathbf{z}_0,\mathbf{z}_1)}
  \bigl[\lVert v_\theta(\mathbf{z}_\tau,\;\tau,\;\mathbf{c}) -
  (\mathbf{z}_1 - \mathbf{z}_0) \rVert^2\bigr],
  \label{eq:lflow}
\end{equation}
where $\mathbb{E}$ denotes the expectation over the flow time $\tau$ and the
sampled training latent pair $(\mathbf{z}_0,\mathbf{z}_1)$.

\paragraph{Local normalized cross-correlation.}
An image-space registration loss between each warped ED volume and the
corresponding ground-truth frame $\mathbf{x}_t^{\mathrm{GT}}$, robust to
intensity offsets:
\begin{equation}
  \mathcal{L}_{\mathrm{lncc}} = 1 - \frac{1}{T} \sum_{t=1}^{T} \mathrm{LNCC}\bigl(
    \mathrm{warp}(\mathbf{x}_{\mathrm{ED}},\,\boldsymbol{\varphi}_t),\;
    \mathbf{x}_t^{\mathrm{GT}}\bigr),
  \label{eq:llncc}
\end{equation}
computed over local windows on each 2D slice of the volume.

\paragraph{Deformation regularization.}
The bending energy~\cite{rueckert1999nonrigid} penalizes second-order spatial
derivatives of the displacement field, extended to 3D:
\begin{equation}
  \mathcal{L}_{\mathrm{bend}} = \sum_{i \in \{x,y,z\}}
    \lVert \partial^2 \boldsymbol{\varphi} / \partial x_i^2 \rVert^2
    + 2\sum_{i \neq j} \lVert \partial^2 \boldsymbol{\varphi} / \partial x_i\,\partial x_j \rVert^2.
  \label{eq:lbend}
\end{equation}
The folding loss penalizes topology violations only, without forcing the
Jacobian determinant toward unity, as this would contradict the physiological
volume change of the LV cavity:
\begin{equation}
  \mathcal{L}_{\mathrm{fold}} =
    \bigl\langle \mathrm{ReLU}(-\det\mathbf{J}_{\boldsymbol{\varphi}})^2 \bigr\rangle,
  \label{eq:lfold}
\end{equation}
where $\langle\cdot\rangle$ denotes the spatial mean, $\mathrm{ReLU}(a)=\max(a,0)$
so that only folded voxels with a negative Jacobian determinant are penalized,
and $\det\mathbf{J}$ is computed via 3D finite differences.

\paragraph{ED identity loss.}
The displacement at the ED frame must vanish:
\begin{equation}
  \mathcal{L}_{\mathrm{ed}} = 1 - \mathrm{Dice}\bigl(
    \mathrm{warp}(\mathbf{m}_{\mathrm{ED}},\boldsymbol{\varphi}_{t_{\mathrm{ED}}}),\,
    \mathbf{m}_{\mathrm{ED}}\bigr)
  + \lVert \boldsymbol{\varphi}_{t_{\mathrm{ED}}} \rVert_1,
  \label{eq:led}
\end{equation}
where $\mathbf{m}_{\mathrm{ED}}$ is the binary LV mask at ED, $\mathrm{Dice}$ is
the soft Dice overlap, and $\lVert\cdot\rVert_1$ is the $L_1$ norm, jointly
penalizing mask misalignment and non-zero displacement at ED.

\paragraph{ES endpoint supervision.}
Explicit supervision at the end-systolic frame anchors the contraction
magnitude:
\begin{equation}
  \mathcal{L}_{\mathrm{es}} = 1 - \mathrm{Dice}^{\mathrm{es}}
  + \lambda_{\mathrm{esv}}\lvert V_{\mathrm{ES}}^{\mathrm{pred}} - V_{\mathrm{ES}}^{\mathrm{GT}}\rvert
  + \lambda_{\mathrm{ef}}\lvert \mathrm{EF}^{\mathrm{pred}} - \mathrm{EF}^{\mathrm{GT}}\rvert,
  \label{eq:les}
\end{equation}
where $1 - \mathrm{Dice}^{\mathrm{es}}$ is the soft Dice term evaluated at
ES, $V_{\mathrm{ES}}$ is the end-systolic LV volume, and $\lambda_{\mathrm{esv}}$,
$\lambda_{\mathrm{ef}}$ are scalar weights.
Predicted volumes and EF are obtained by warping the ground-truth ED
segmentation mask through the predicted displacement field, so that no
segmentation network ever runs on generated images.

\paragraph{Volume curve loss.}
Dense temporal supervision over the full LV volume trajectory:
\begin{equation}
  \mathcal{L}_{\mathrm{vol}} = 0.5\,\mathcal{L}_{\mathrm{MAE}}
    + 0.5\,(1 - R^2)
    + 0.5\,\mathcal{L}_{\mathrm{amp}}
    + 0.2\,\mathcal{L}_{\mathrm{bias}},
  \label{eq:lvol}
\end{equation}
where $\mathcal{L}_{\mathrm{MAE}}$ is the per-frame normalized volume error,
$R^2$ is computed between the predicted and ground-truth volume curves,
$\mathcal{L}_{\mathrm{amp}}$ penalizes amplitude mismatch, and
$\mathcal{L}_{\mathrm{bias}}$ penalizes systematic volume offset.
The predicted volume at frame $t$ is obtained by warping the frozen SegUNet's
soft ED mask through $\boldsymbol{\varphi}_t$; ground-truth volumes come from
SegUNet applied to real frames.
During training, the frozen SegUNet thus processes only real CMR images, so
domain shift is avoided
and gradients remain reliable, flowing through the differentiable warp back to
the velocity network.

Therefore, the total loss combines flow-matching, image-space, and physiological terms:
\begin{align}
  \mathcal{L} &= \lambda_{\mathrm{flow}}\,\mathcal{L}_{\mathrm{flow}}
  + \lambda_{\mathrm{lncc}}\,\mathcal{L}_{\mathrm{lncc}}
  + \lambda_{\mathrm{bend}}\,\mathcal{L}_{\mathrm{bend}}
  + \lambda_{\mathrm{fold}}\,\mathcal{L}_{\mathrm{fold}}
  + \lambda_{\mathrm{ed}}\,\mathcal{L}_{\mathrm{ed}}
  + \lambda_{\mathrm{es}}\,\mathcal{L}_{\mathrm{es}}
  + \lambda_{\mathrm{vol}}\,\mathcal{L}_{\mathrm{vol}}
  + \lambda_{\mathrm{res}}\,\mathcal{L}_{\mathrm{res}}.
  \label{eq:total_loss}
\end{align}
where each $\lambda$ is a scalar weight balancing the corresponding term.

\subsection{Network architecture and implementation}
\label{sec:architecture_impl}

\begin{figure*}[width=\textwidth,pos=t]
  \centering
  \includegraphics[width=\linewidth]{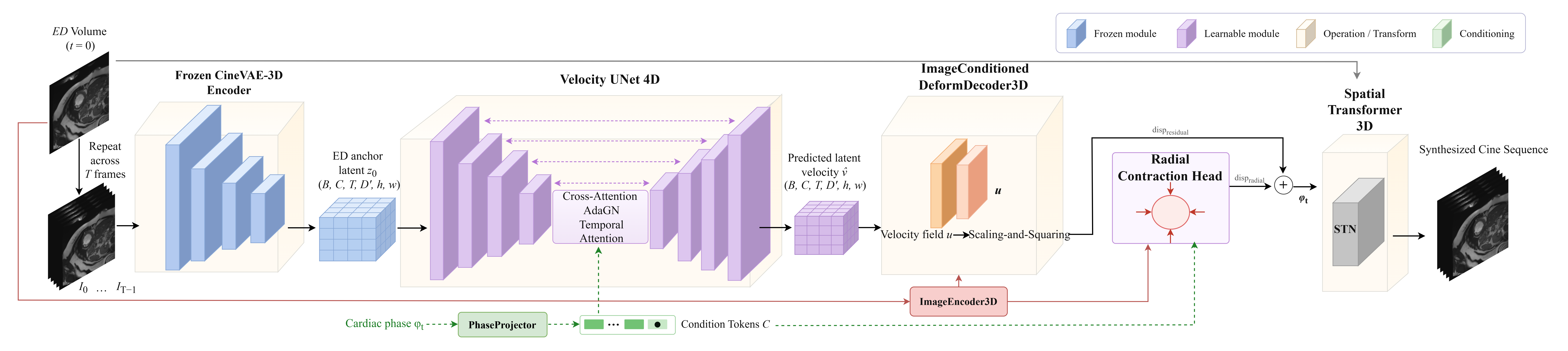}
  \caption{\textbf{PhaseFlow3D} encodes the ED 3D volume into a latent anchor
    $\mathbf{z}_0$, predicts a phase-conditioned latent velocity via rectified
    flow matching, and decodes a 3D displacement field through
    a Radial Contraction Decomposition. The displacement field directly warps
    the ED volume to synthesize the full 4D cine sequence.}
  \label{fig:architecture}
\end{figure*}

As Figure~\ref{fig:architecture} illustrates, PhaseFlow3D consists of six
components that transport the single ED volume into the full 4D cine sequence:

\begin{itemize}
  \item CineVAE3D encoder ($E_{\mathrm{vae}}$, frozen):
    a per-frame 3D VAE that maps each frame $(1, D, H, W)$ to a latent
    $(4, D', h, w) = (4, 4, 16, 16)$.
    It encodes the ED volume into the anchor latent $\mathbf{z}_0$
    (Eq.~\eqref{eq:z0}), using 3D convolutions with per-frame 2D spatial
    attention at the bottleneck.
  \item PhaseProjector:
    a small MLP that maps each phase code
    $\mathbf{p}_t = [\sin\phi_t, \cos\phi_t]$ to a per-frame embedding
    $\mathbf{e}_{\mathrm{phase}} \in \mathbb{R}^{T \times d}$.
    The embedding is injected at every resolution level through
    cross-attention.
  \item VelocityUNet4D ($v_\theta$):
    a 3D U-Net (base channels $128$, multipliers $[1,2]$) that predicts the
    latent velocity $\hat{\mathbf{v}}$ (Eq.~\eqref{eq:velocity}).
    It combines temporal attention across frames, per-frame spatial attention,
    AdaGN residual blocks, and phase cross-attention
    ($n_{\mathrm{tokens}}=4$, $n_{\mathrm{heads}}=8$).
    The D$'$-fold trick folds the depth axis into channels, so full 4D
    reasoning runs at the cost of a 3D network.
  \item RadialContractionHead:
    the physics-informed head that produces the radial displacement
    $\boldsymbol{\varphi}_t^{\mathrm{radial}}$ (Eq.~\eqref{eq:radial}).
    A GlobalNet pools the ED features into the patient amplitude
    $A_{\mathrm{pat}}$, a LocalNet outputs the spatial modulation map
    $\mathbf{m}_{\mathrm{loc}}$, and a TemporalNet outputs the activation curve
    $s_t$.
  \item ImageEncoder3D and ImageConditionedDeformDecoder3D:
    the data-driven pathway that decodes the residual displacement
    $\boldsymbol{\varphi}^{\mathrm{residual}}$ (Eq.~\eqref{eq:residual}).
    A lightweight ImageEncoder3D (channels $[16, 32, 64]$) supplies multi-scale
    skip features of the ED volume to the transposed-convolution decoder, and a
    ScalingSquaring3D layer ($K=7$) integrates the residual velocity field.
  \item SpatialTransformer3D:
    a differentiable warp with trilinear sampling that deforms the ED volume by
    the total field
    $\boldsymbol{\varphi}_t = \boldsymbol{\varphi}_t^{\mathrm{radial}}
    + \boldsymbol{\varphi}_t^{\mathrm{residual}}$ to synthesize each frame
    (Eq.~\eqref{eq:warp}).
\end{itemize}

The full framework is implemented in Python and PyTorch.
A frozen 2D SegUNet supplies segmentation masks for the physiological losses
during training only, and never processes generated images.
Complete layer specifications and hyperparameters are given in
Appendix~\ref{app:implementation}.

\section{Experiments and Results}
\label{sec:experiments}

\subsection{Experiment Setup}
\label{sec:setup}

\paragraph{Datasets.}
We trained and evaluated on two publicly available cardiac CMR datasets, which are ACDC and M\&Ms.
ACDC~\cite{bernard2018acdc} contains 150 subjects from five pathological groups: normal control (NOR), dilated cardiomyopathy (DCM), hypertrophic cardiomyopathy (HCM), prior myocardial infarction (MINF), and abnormal right ventricle (ARV), with 30 subjects per group. Each subject includes a cine CMR sequence with ED/ES annotation. We split the dataset into 105 training subjects, 15 validation subjects, and 30 test subjects in a stratified manner, with the five pathological groups evenly distributed across all splits.
M\&Ms~\cite{campello2021mms} is the Multi-Centre,
Multi-Vendor \& Multi-Disease dataset acquired on four scanner vendors across
multiple clinical centres; we retained the challenge split assignments for
the included cases, using $150$ labeled training, $34$ validation, and $136$
testing subjects.
The model was trained jointly on the ACDC and M\&Ms training subjects
($255$ subjects in total) and evaluated on the combined ACDC and M\&Ms
test sets.
All volumes were resampled to $D=16$ slices,
$H=W=128$ in-plane, $T=30$ frames per cycle, and normalized to $[-1,1]$.

\paragraph{Implementation.}
CineVAE3D was pretrained for 100 epochs with a learning rate of $10^{-4}$ and KL weight of $10^{-6}$,
and frozen during flow training.
The frozen SegUNet is a standard 2D U-Net~\cite{ronneberger2015unet}
trained on ACDC segmentation annotations for the LV, right ventricle (RV), and
myocardium, used for ED/ES mask extraction and volume-curve computation;
during training it was never exposed to generated images.
The network architecture and its layer dimensions are specified in
Sec.~\ref{sec:architecture_impl}, with full details in
Appendix~\ref{app:implementation}.
At inference, the canonical phase template with population-average ES fraction
$\bar{r}_{\mathrm{ES}} = 0.38$ was used (Sec.~\ref{sec:phase}).

\paragraph{Training.}
PhaseFlow3D was trained in two progressive stages. The full two-stage schedule took approximately 29 hours on a single NVIDIA RTX A6000 GPU.
Jointly optimizing all components from scratch is unstable, because the
flow-matching network can learn spurious shortcuts before the radial head has
established a correct contraction prior.
In Stage~1 (50 epochs), only the RadialContractionHead and PhaseProjector were
trainable, while the velocity network and deformation decoder stayed frozen.
The loss was restricted to the physiology terms
($\mathcal{L}_{\mathrm{es}} + \mathcal{L}_{\mathrm{ed}}
+ \mathcal{L}_{\mathrm{vol}} + \mathcal{L}_{\mathrm{bend}}
+ \mathcal{L}_{\mathrm{fold}}$), so the radial head learned the correct
contraction direction and amplitude before any image-level gradient biased the
temporal curve.
In Stage~2 (100 epochs), the VelocityUNet4D, ImageEncoder3D, and deformation
decoder were unfrozen, and all terms of Eq.~\eqref{eq:total_loss} were active with
physiology-biased weights.
The best checkpoint (epoch 60) was used in all experiments.
Both stages used AdamW~\cite{loshchilov2018adamw} with $\beta_1$ of $0.9$,
$\beta_2$ of $0.999$ and weight decay of $10^{-4}$, learning rate $10^{-4}$ with cosine
annealing to $10^{-6}$, batch size 1, and mixed-precision training.

\paragraph{Evaluation metrics.}
We organized evaluation into two categories:
\begin{itemize}
  \item Image and video quality: Peak Signal-to-Noise Ratio~(PSNR), Structural Similarity Index Measure~\cite{wang2004ssim}~(SSIM), Learned Perceptual Image Patch Similarity~\cite{zhang2018lpips}~(LPIPS), Fr\'echet Inception Distance~\cite{heusel2017fid}~(FID), and Fr\'echet Video Distance (FVD)~\cite{unterthiner2019fvd}, computed on full synthesized sequences to measure temporal distributional quality.
  \item Physiological fidelity: Ejection Fraction Mean Absolute Error~(EF MAE),
    Volume Curve Pearson Correlation~(Vol Corr), Volume Coefficient of Determination~(Vol~$R^2$), Volume Mean Absolute Error~(Vol~MAE).
\end{itemize}

Formal definitions and the motivation for each metric are given in
Appendix~\ref{app:metrics}.

\subsection{Comparison to Baselines}

In this experiment, we evaluated the model performance on the combined ACDC and
M\&Ms test set, which spans five pathology groups and multiple scanner vendors.
Table~\ref{tab:main} shows the performance of the proposed model (PhaseFlow3D)
compared with seven other state-of-the-art techniques on the test sequences in
terms of image and video quality (PSNR, SSIM, LPIPS, FID, and FVD) and
physiological fidelity (EF MAE, Vol Corr, Vol~$R^2$, and Vol MAE). All seven
baselines were retrained on this split under identical data conditions, while
ECG-conditioned methods~\cite{echopulse, ecgflowcmr} and registration
methods~\cite{voxelmorph} were excluded, as the former require an ECG input
that violates our ECG-free setting and the latter require a target
frame at inference rather than synthesizing from a single ED volume. DragNet, a
published method in the ECG-free, single-frame setting, was reproduced
in its native 2D modality because our whole-volume 3D adaptation
collapsed the predicted displacement toward zero when the static
background dominates the reconstruction loss; we therefore ran the unmodified 2D
model one short-axis slice at a time and resampled the stacked volumes to the
shared frame count $T$, placing it in the same evaluation harness as every other
method under a like-for-like comparison.

\begin{table*}[pos=htbp]
\centering
\small
\setlength{\tabcolsep}{3.0pt}
\caption{Main results on the combined ACDC and M\&Ms test set.
Best in \textbf{bold}, second-best \underline{underlined}.
Baselines are evaluated with image-based EF (SegUNet on generated frames),
whereas PhaseFlow3D uses warped ED masks; EF under unified measurement
protocols is cross-validated in Appendix~\ref{app:metrics}
(Table~\ref{tab:ef_protocols}); ``---'' denotes inapplicable metrics.}
\label{tab:main}
\begin{tabular}{l ccccc cccc}
\toprule
& \multicolumn{5}{c}{\textbf{Image \& Video Quality}} &
  \multicolumn{4}{c}{\textbf{Physiological Fidelity}} \\
\cmidrule(lr){2-6} \cmidrule(lr){7-10}
Method
  & PSNR$\uparrow$ & SSIM$\uparrow$ & LPIPS$\downarrow$ & FID$\downarrow$ & FVD$\downarrow$
  & EF MAE$\downarrow$ & Vol Corr$\uparrow$ & Vol $R^2$$\uparrow$ & Vol MAE$\downarrow$ \\
\midrule
ConvLSTM~\cite{shi2015convlstm}
  & 19.16 & 0.426 & 0.474 & 292.0 & 23.31
  & --- & --- & $-$37.80 & 0.734 \\
GAN-cMRI~\cite{gancmri}
  & 22.64 & 0.542 & 0.409 & 185.8 & 19.70
  & 39.35\% & 0.308 & $-$32.91 & 0.673 \\
LFDM~\cite{lfdm}
  & 28.50 & 0.800 & 0.200 & 82.53 & 11.24
  & 29.83\% & \underline{0.809} & $-$5.547 & 0.221 \\
CVAE
  & 33.01 & 0.899 & 0.066 & 27.95 & 4.141
  & 32.39\% & 0.781 & $-$0.582 & 0.179 \\
DragNet~\cite{dragnet}
  & 35.61 & 0.928 & 0.030 & 5.450 & 1.690
  & 37.77\% & 0.757 & $-$0.484 & \underline{0.175} \\
SimVP~\cite{simvp}
  & \underline{35.79} & 0.920 & 0.023 & 6.080 & 1.700
  & \underline{27.03}\% & 0.644 & $-$0.711 & 0.187 \\
Direct Reg
  & \textbf{36.37} & \underline{0.928} & \underline{0.020} & \underline{3.374} & \underline{1.042}
  & 30.14\% & 0.590 & $-$0.892 & 0.194 \\
\midrule
\textbf{PhaseFlow3D (Ours)}
  & 34.97 & \textbf{0.928} & \textbf{0.017} & \textbf{2.020} & \textbf{0.683}
  & \textbf{12.85\%} & \textbf{0.845} & \textbf{0.054} & \textbf{0.110} \\
\bottomrule
\end{tabular}
\end{table*}

The results indicate that PhaseFlow3D reproduces cardiac physiology far more
faithfully than the competing methods. Its EF MAE of $12.85\%$ roughly halved
that of the best competing method ($27.03\%$), and it was the only model to
reach a positive Vol~$R^2$ ($0.05$); every baseline stayed strongly negative,
from $-0.58$ down to $-37.80$, which reveals a systematic failure to recover the
ventricular volume trajectory. Even the variant that regresses the frame
intensities directly from the same ED volume and phase input incurred more
than twice our EF error ($30.14\%$), indicating that the gain came from
explicit physiological grounding rather than from the phase conditioning
alone.

This advantage carried over to whole-sequence quality. PhaseFlow3D achieved the
best FID ($2.02$) and FVD ($0.68$), improving video-level coherence over the
strongest baseline ($1.04$) by roughly one third, even though a few methods
reached marginally higher per-frame PSNR ($36.37$~dB against our $34.97$~dB).
The different PSNR and distributional-quality rankings are consistent with
the distinction between distortion and perceptual distribution matching
discussed by Blau and Michaeli~\cite{blau2018perception}. The EF and LV
volume-curve differences are separate empirical findings of our cardiac
experiments, rather than consequences of that theory.

\paragraph{Statistical significance.}
To confirm that these physiological improvements reflect genuine gains rather
than point-estimate artifacts, we applied the paired Wilcoxon signed-rank test
on the per-patient physiological metrics ($n{=}166$, pairing by patient)
against all seven baselines and the trivial static null (ED~Repeat).
Table~\ref{tab:wilcoxon} reports the resulting $p$-values; each cell is a
$p$-value rather than a performance score, so a smaller value is stronger
evidence in favour of PhaseFlow3D, and in every significant cell the per-patient
means favour PhaseFlow3D.
PhaseFlow3D was significantly better than every baseline on EF~MAE, Vol~$R^2$,
and Vol~MAE, with $p$-values ranging from $10^{-11}$ down to $10^{-29}$.
The single non-significant cell was Vol~Corr against LFDM ($p{=}0.84$): LFDM
tracks when the ventricle contracts but, as its negative $R^2$ ($-5.55$) shows,
not by how much, which is precisely the failure mode the volume-curve
supervision is designed to correct.
Three cells are undefined and omitted: Vol~Corr for ED~Repeat, whose constant
zero-variance volume curve leaves Pearson correlation undefined, and
EF~MAE/Vol~Corr for ConvLSTM, whose generated frames contain no LV cavity for
the segmentation network to detect, matching the dashes in
Table~\ref{tab:main}; for GAN-cMRI, EF~MAE and Vol~Corr are computed over the
$n{=}128$ patients with a detectable LV cavity.

\begin{figure*}[width=\textwidth,pos=t]
  \centering
  \includegraphics[width=\linewidth]{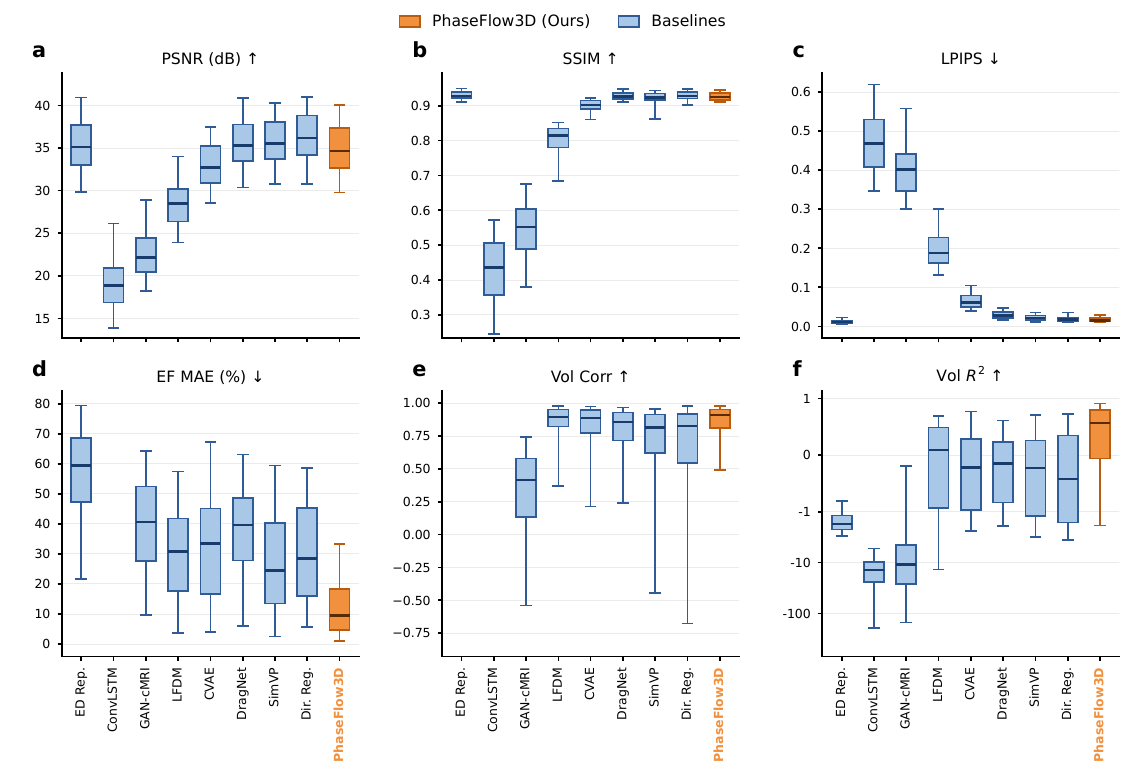}
  \caption{Per-patient distribution of image quality (PSNR, SSIM, LPIPS) and
    physiological fidelity (EF MAE, Vol~Corr, Vol~$R^2$) on the combined ACDC
    and M\&Ms test set ($n{=}166$). Each box spans the interquartile range with
    the median line and 5--95th percentile whiskers. PhaseFlow3D (orange) is
    compared against the eight baselines (blue); empty slots mark metrics that
    are undefined for a given method, matching the dashes in
    Table~\ref{tab:main}. Vol~$R^2$ uses a symmetric-log axis to resolve both
    the high-performing methods near one and the collapsed baselines.}
  \label{fig:boxplots}
\end{figure*}

While Table~\ref{tab:main} reports cohort means, Figure~\ref{fig:boxplots} shows
the full per-patient distribution of every metric across all 166 test subjects.
PhaseFlow3D attained the highest median and the tightest spread, confirming that
its advantage is consistent rather than driven by a few easy cases.

\begin{table*}[pos=htbp]
  \centering
  \small
  \setlength{\tabcolsep}{5pt}
  \caption{Paired Wilcoxon signed-rank test ($n{=}166$ patients) against all
    baselines. Every cell is a $p$-value; smaller means stronger evidence
    that PhaseFlow3D is better, and in every significant cell the
    per-patient means favour PhaseFlow3D. \textbf{Bold} $=$ significant at
    $\alpha{=}0.05$. The only non-significant cell is Vol~Corr vs.\ LFDM;
    ``---'' marks undefined comparisons (constant or empty predicted volume
    curves).}
  \label{tab:wilcoxon}
  \begin{tabular}{lcccccccc}
    \toprule
    & \multicolumn{8}{c}{Wilcoxon $p$ (PhaseFlow3D vs.)} \\
    \cmidrule(lr){2-9}
    Metric & ED~Rep. & ConvLSTM & GAN-cMRI & LFDM & CVAE & DragNet & SimVP & Dir.~Reg. \\
    \midrule
    EF MAE    & $\mathbf{4.5{\times}10^{-28}}$ & ---                            & $\mathbf{9.5{\times}10^{-20}}$ & $\mathbf{2.5{\times}10^{-18}}$ & $\mathbf{1.1{\times}10^{-19}}$ & $\mathbf{9.8{\times}10^{-23}}$ & $\mathbf{1.1{\times}10^{-14}}$ & $\mathbf{1.6{\times}10^{-17}}$ \\
    Vol Corr  & ---                            & ---                            & $\mathbf{2.2{\times}10^{-22}}$ & $0.84$                         & $\mathbf{5.6{\times}10^{-3}}$  & $\mathbf{1.0{\times}10^{-2}}$  & $\mathbf{1.1{\times}10^{-10}}$ & $\mathbf{3.9{\times}10^{-9}}$ \\
    Vol $R^2$ & $\mathbf{1.3{\times}10^{-21}}$ & $\mathbf{5.4{\times}10^{-29}}$ & $\mathbf{8.4{\times}10^{-29}}$ & $\mathbf{4.7{\times}10^{-13}}$ & $\mathbf{6.7{\times}10^{-11}}$ & $\mathbf{6.1{\times}10^{-10}}$ & $\mathbf{2.5{\times}10^{-12}}$ & $\mathbf{1.3{\times}10^{-13}}$ \\
    Vol MAE   & $\mathbf{2.2{\times}10^{-27}}$ & $\mathbf{5.4{\times}10^{-29}}$ & $\mathbf{7.9{\times}10^{-29}}$ & $\mathbf{3.7{\times}10^{-18}}$ & $\mathbf{1.1{\times}10^{-17}}$ & $\mathbf{2.8{\times}10^{-16}}$ & $\mathbf{2.5{\times}10^{-19}}$ & $\mathbf{5.0{\times}10^{-19}}$ \\
    \bottomrule
  \end{tabular}
\end{table*}

\subsection{Ablation Study}
\label{sec:ablation}

To quantify the contribution of each proposed component, we removed them one at
a time and retrained the model under otherwise identical settings.
Table~\ref{tab:ablation} reports the outcome: every removal produced a distinct
failure mode rather than a uniform degradation, confirming that the components
are complementary rather than redundant.

\begin{table*}[pos=htbp]
\centering
\small
\setlength{\tabcolsep}{3.0pt}
\caption{Ablation study on the combined ACDC and M\&Ms test set.
Best in \textbf{bold}, second-best \underline{underlined}.}
\label{tab:ablation}
\begin{tabular}{l ccccc cccc}
\toprule
& \multicolumn{5}{c}{\textbf{Image \& Video Quality}} &
  \multicolumn{4}{c}{\textbf{Physiological Fidelity}} \\
\cmidrule(lr){2-6} \cmidrule(lr){7-10}
Variant
  & PSNR$\uparrow$ & SSIM$\uparrow$ & LPIPS$\downarrow$ & FID$\downarrow$ & FVD$\downarrow$
  & EF MAE$\downarrow$ & Vol Corr$\uparrow$ & Vol $R^2$$\uparrow$ & Vol MAE$\downarrow$ \\
\midrule
A1: w/o deform decoder
  & \textbf{35.37} & \textbf{0.929} & \textbf{0.012} & \underline{1.949} & 0.724
  & 55.33\% & --- & $-$1.766 & 0.250 \\
A2: w/o radial head
  & 34.48 & 0.924 & 0.017 & 2.155 & 0.765
  & 15.00\% & \textbf{0.851} & $-$0.244 & \underline{0.114} \\
A3: radial only (w/o residual)
  & 33.14 & 0.922 & 0.020 & 3.098 & 1.058
  & 18.48\% & 0.823 & $-$0.415 & 0.132 \\
A4: w/o vol curve loss
  & 34.33 & 0.924 & 0.020 & 3.063 & 0.998
  & 14.66\% & 0.785 & $-$0.643 & 0.143 \\
A5: linear phase
  & 34.91 & \underline{0.927} & \underline{0.017} & \textbf{1.853} & \underline{0.697}
  & \textbf{12.60\%} & 0.741 & $-$0.198 & 0.126 \\
\midrule
Full model
  & \underline{34.97} & \underline{0.927} & \underline{0.017} & 2.020 & \textbf{0.683}
  & \underline{12.85\%} & \underline{0.845} & \textbf{0.054} & \textbf{0.110} \\
\bottomrule
\end{tabular}
\end{table*}

\paragraph{A1: w/o deformation decoder.}
Without the deformation pipeline, synthesis falls back to pure VAE decoding,
$\hat{\mathbf{x}}_t = D_{\mathrm{vae}}(\mathbf{z}_0 + \hat{\mathbf{v}})^{(t)}$.
Per-frame metrics remain nominally strong (PSNR $35.37$~dB, LPIPS $0.012$),
yet EF MAE collapses to $55.33\%$ and Vol~Corr becomes undefined: the
generated frames contain essentially no ventricular contraction, equivalent to
predicting a static ED volume.
Image-based evaluation (running SegUNet on the generated frames) likewise
yields ${\sim}55\%$ EF MAE, confirming that no cardiac motion is produced.
The displacement field is thus the sole mechanism through which cardiac motion
is encoded, and the prerequisite for applying any physiological loss.

\paragraph{A2: w/o radial head.}
Leaving all displacement to the data-driven residual degrades EF MAE from
$12.85\%$ to $15.00\%$ and inverts Vol~$R^2$ from $0.05$ to $-0.24$.
The physics-informed prior contributes accuracy that the residual cannot
recover on its own: without the radial prior, contraction amplitude
and its temporal curve become unreliable.

\paragraph{A3: radial only.}
Conversely, removing the residual costs image quality: PSNR drops to
$33.14$~dB and FVD rises to $1.06$.
A purely radial field cannot represent rotation, twist, or out-of-plane
motion.
The two components are complementary: the radial part carries the physics, the
residual carries everything else.

\paragraph{A4: w/o volume curve loss.}
Image quality is largely preserved (PSNR $34.33$~dB) but EF MAE rises to
$14.66\%$ and Vol~Corr falls to $0.78$: the ES endpoint terms alone anchor two
points of the trajectory but cannot constrain its full temporal course.
Dense volume-curve supervision is necessary for the complete
systolic--diastolic cycle.

\paragraph{A5: linear phase.}
Replacing the piecewise phase with uniform $\phi_t^{\mathrm{lin}} = 2\pi t/T$
leaves EF MAE essentially unchanged ($12.60\%$ vs.\ $12.85\%$) but
substantially degrades Vol~Corr ($0.74$ vs.\ $0.84$) and Vol~$R^2$ ($-0.20$
vs.\ $0.05$).
The piecewise phase therefore contributes chiefly to \emph{volume-curve
shape}: it lets the conditioning distinguish the rapid systolic descent from
the slower diastolic recovery, reproducing the asymmetric trajectory that a
uniform encoding cannot.

\subsection{Phase Template Validation}
\label{sec:phase_validation}

The canonical phase template (Sec.~\ref{sec:phase}) replaces patient-specific
ES timing with a single population-average ES fraction
$\bar{r}_{\mathrm{ES}} = 0.38$.
We quantified the approximation this introduces, using the individual ED and ES
timings of the $255$ training subjects.

\begin{figure}[pos=t]
  \centering
  \includegraphics[width=0.6\linewidth]{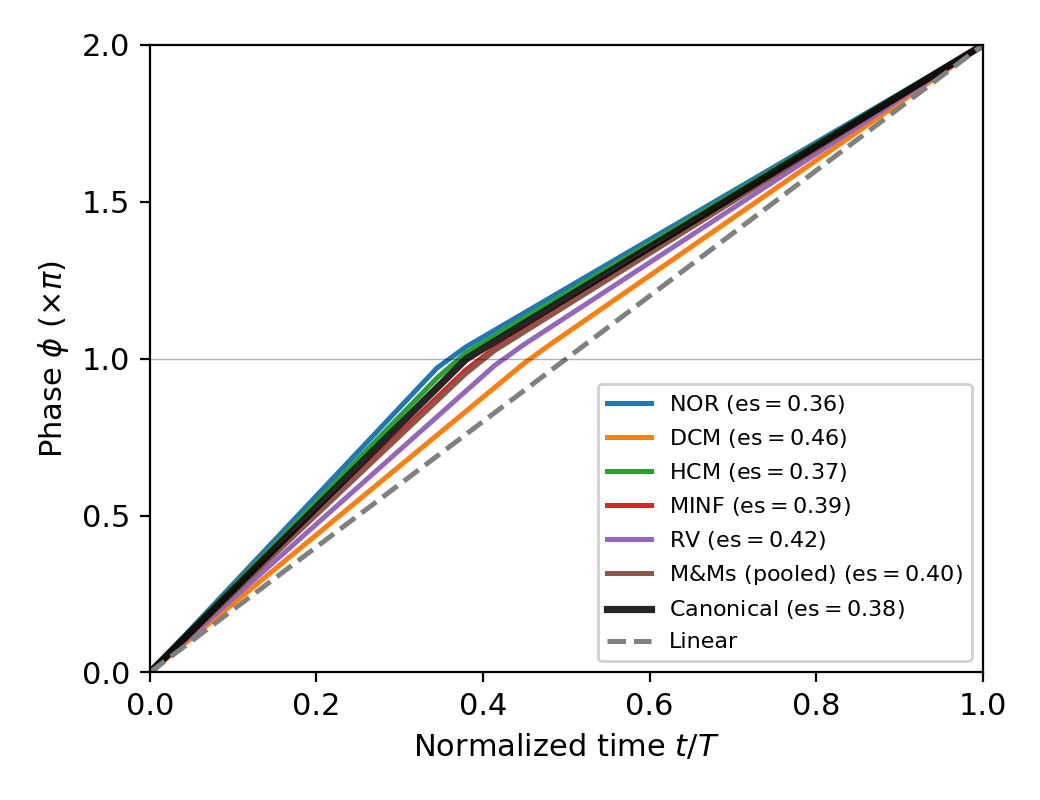}
  \caption{Group-mean piecewise phase templates (coloured; ACDC pathology
    groups and pooled M\&Ms) vs.\ the canonical template used at inference
    (black) and the uniform linear baseline (grey dashed). Each template
    reaches $\phi = \pi$ (ES) at the group-mean ES fraction, encoding
    pathology-specific systole--diastole timing.}
  \label{fig:phase_templates}
\end{figure}

\paragraph{Empirical phase templates.}
Figure~\ref{fig:phase_templates} shows the mean piecewise phase templates of
each ACDC pathology group and of the pooled M\&Ms cohort, together with the
uniform linear baseline and the canonical template
($\bar{r}_{\mathrm{ES}}=0.38$) used at blind inference.
The empirical ES fractions differed systematically across groups: NOR subjects
completed systole earliest ($0.355 \pm 0.063$), followed by HCM
($0.367 \pm 0.054$) and MINF ($0.393 \pm 0.079$), whereas ARV
($0.423 \pm 0.127$) and in particular DCM ($0.456 \pm 0.088$) exhibited a
prolonged systole, consistent with impaired contractility in dilated
cardiomyopathy.
The pooled M\&Ms cohort ($0.398 \pm 0.060$) and the overall training
population ($0.398 \pm 0.076$, $n=255$) were both well approximated by the
canonical value of $0.38$, which lay within a third of a standard deviation of
the population mean.

\begin{figure*}[width=\textwidth,pos=t]
  \centering
  \includegraphics[width=\linewidth]{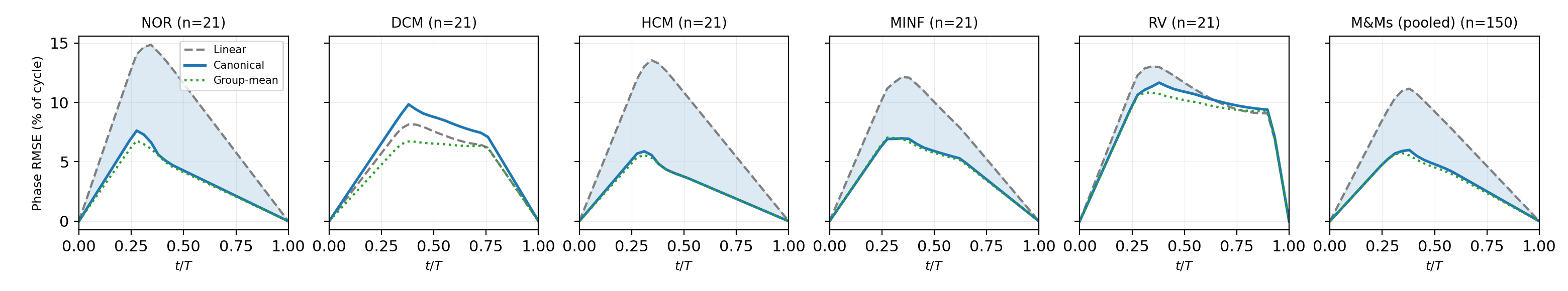}
  \caption{Per-frame phase RMSE (\% of cycle) of the canonical template
    (solid blue), the group-mean oracle template (dotted green), and the
    uniform linear baseline (grey dashed), per ACDC pathology group and for
    the pooled M\&Ms cohort. Shaded region: error saved by the canonical
    template over the linear baseline.}
  \label{fig:intra_group_consistency}
\end{figure*}

\paragraph{Intra-group consistency.}
To quantify how well a population-level template approximates individual patient
phase curves, we measured the per-frame root mean squared error (RMSE) of each
training subject's piecewise phase $\phi_t(r_{\mathrm{ES}}^{(i)})$ relative to
three candidates: the canonical template $\phi_t(0.38)$ used at blind
inference; the group-mean template $\phi_t(\bar{r}_{\mathrm{ES},g})$, an
oracle that requires the pathology label; and the uniform linear phase
$\phi_t^{\text{lin}} = 2\pi t / T$.
Pooled over all $255$ training subjects, the canonical template achieved a mean
RMSE of $3.8\%$ of the full cardiac cycle against $6.2\%$ for the linear phase,
a $1.6\times$ overall reduction; the largest gains occurred in HCM
($2.71\%$ vs.\ $6.93\%$, $2.6\times$) and NOR ($3.31\%$ vs.\ $7.65\%$,
$2.3\times$), with M\&Ms at $3.10\%$ vs.\ $5.74\%$ ($1.9\times$).
The two exceptions were instructive: for ARV the two templates were nearly
equivalent ($8.21\%$ vs.\ $8.69\%$), and for DCM, whose delayed ES timing
($0.456$) happens to lie close to the linear midpoint, the linear phase was
marginally closer ($4.93\%$ vs.\ $5.72\%$).
The group-mean oracle improved only slightly over the canonical template in
every group (Figure~\ref{fig:intra_group_consistency}), indicating that a single
population-level template captures most of the attainable benefit without
requiring any pathology label at inference.

\subsection{Deformation Regularity Evaluation}
\label{sec:diffeomorphism}

Scaling-and-squaring parameterizes the residual through velocity integration,
while the folding penalty discourages negative Jacobian determinants in the
total transformation (Sec.~\ref{sec:loss}). We evaluate the total field
empirically because its additive construction does not guarantee a
fold-free transformation.
The Jacobian of the coordinate transformation $\mathrm{Id}+\boldsymbol{\varphi}$
is a $3{\times}3$ matrix,
\begin{equation}
  \mathbf{J}_{\boldsymbol{\varphi}} =
  \mathbf{I}_3 +
  \begin{pmatrix}
    \tfrac{\partial\varphi_x}{\partial x} &
    \tfrac{\partial\varphi_x}{\partial y} &
    \tfrac{\partial\varphi_x}{\partial z} \\[4pt]
    \tfrac{\partial\varphi_y}{\partial x} &
    \tfrac{\partial\varphi_y}{\partial y} &
    \tfrac{\partial\varphi_y}{\partial z} \\[4pt]
    \tfrac{\partial\varphi_z}{\partial x} &
    \tfrac{\partial\varphi_z}{\partial y} &
    \tfrac{\partial\varphi_z}{\partial z}
  \end{pmatrix},
  \label{eq:jacobian}
\end{equation}
which we computed via central finite differences at every voxel of every
generated volume on the test set.

\begin{figure}[pos=t]
  \centering
  \includegraphics[width=\linewidth]{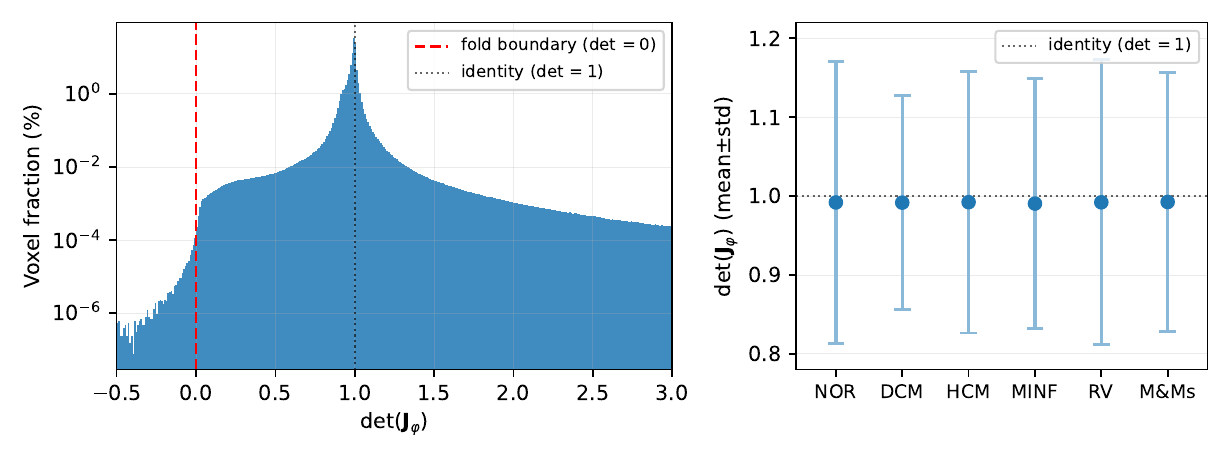}
  \caption{$\det(\mathbf{J}_\varphi)$ distribution of the composed 3D
    displacement field (radial and scaling-and-squaring residual, $K=7$) on
    the ACDC$+$M\&Ms test set.
    \emph{Left}: global histogram (log scale); red dashed: fold boundary
    ($\det=0$).
    \emph{Right}: per-group mean$\pm$std.
    Folding voxels are $0.0005\%$ on average ($<0.002\%$ per patient),
    indicating a low measured folding rate.}
  \label{fig:jacobian_det}
\end{figure}

Figure~\ref{fig:jacobian_det} shows the empirical distribution of
$\det(\mathbf{J}_{\boldsymbol{\varphi}})$ over all $166$ test patients
($30$ frames $\times$ $16{\times}128{\times}128$ voxels each,
${\approx}1.2{\times}10^{9}$ evaluations in total).
The fraction of voxels with $\det(\mathbf{J}_{\boldsymbol{\varphi}})\leq 0$
(folding voxels) was $0.0005\%$ on average and below $0.002\%$ for every
individual patient; topology violations were confined to a few isolated voxels
per volume and never formed coherent folded regions.
The mean determinant was $0.992\pm0.165$. Determinants below and above unity
indicate local contraction and expansion, respectively; these whole-volume
statistics do not isolate myocardial tissue or establish tissue incompressibility.
Per-group statistics (Figure~\ref{fig:jacobian_det}, right) were consistent
across the five ACDC pathology groups and the M\&Ms cohort
($\det(\mathbf{J}_{\boldsymbol{\varphi}})$ between $0.990$ and $0.992$ in every
group), supporting similar measured deformation regularity across the
evaluated cardiac morphologies and acquisition sites.

\begin{figure*}[width=\textwidth,pos=!t]
  \centering
  \includegraphics[width=\linewidth]{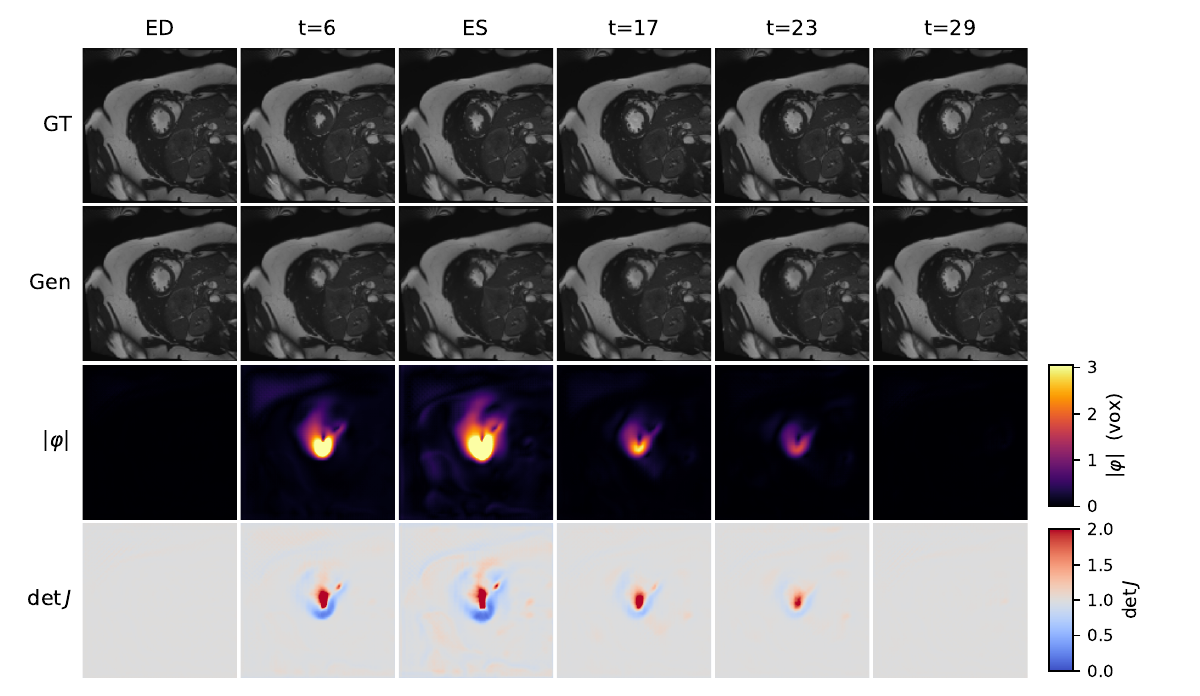}
  \caption{Spatial view of deformation regularity for a representative HCM
    subject (patient036, mid-ventricular slice) at six frames across the cardiac
    cycle. Rows: ground-truth cine, PhaseFlow3D generation, displacement
    magnitude $|\boldsymbol{\varphi}|$, and Jacobian determinant
    $\det(\mathbf{J}_{\boldsymbol{\varphi}})$. The deformation is localized to
    the myocardium; $\det(\mathbf{J}_{\boldsymbol{\varphi}})<1$ (blue) marks
    local contraction and $>1$ (red) local expansion, and it stays strictly
    positive everywhere, confirming a fold-free diffeomorphic mapping.}
  \label{fig:defgrid}
\end{figure*}

Figure~\ref{fig:defgrid} shows the spatial counterpart for a representative HCM
subject: the displacement magnitude $|\boldsymbol{\varphi}|$ stays localized to
the myocardium, and $\det(\mathbf{J}_{\boldsymbol{\varphi}})$ falls below one in
the contracting wall while remaining strictly positive throughout, so the
recovered deformation is both physiologically directed and free of folding.

\subsection{Qualitative Results}
\label{sec:qualitative}

\begin{figure*}[width=\textwidth,pos=!t]
  \centering
  \includegraphics[width=\linewidth]{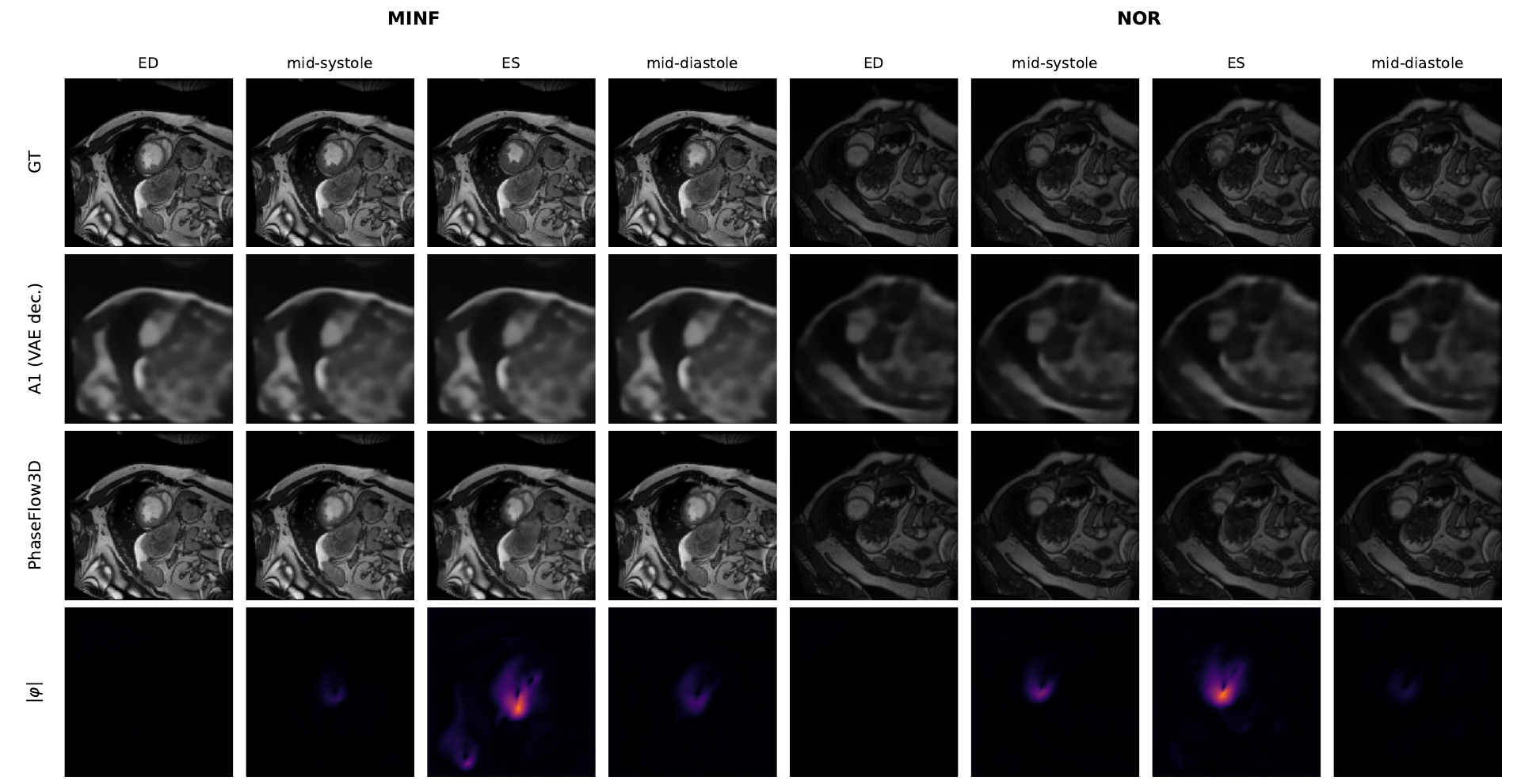}
  \caption{Qualitative comparison (MINF and NOR, mid-ventricular
    slice). Rows 1--3: ground truth, the VAE-decoded variant (A1), and
    PhaseFlow3D at ED, mid-systole, ES, and mid-diastole. Row 4: predicted
    displacement magnitude $|\boldsymbol{\varphi}|$.
    PhaseFlow3D produces sharp, anatomically coherent frames; the VAE-decoded
    variant exhibits characteristic reconstruction blur.}
  \label{fig:qualitative}
\end{figure*}

\begin{figure*}[width=\textwidth,pos=!t]
  \centering
  \includegraphics[width=0.85\linewidth]{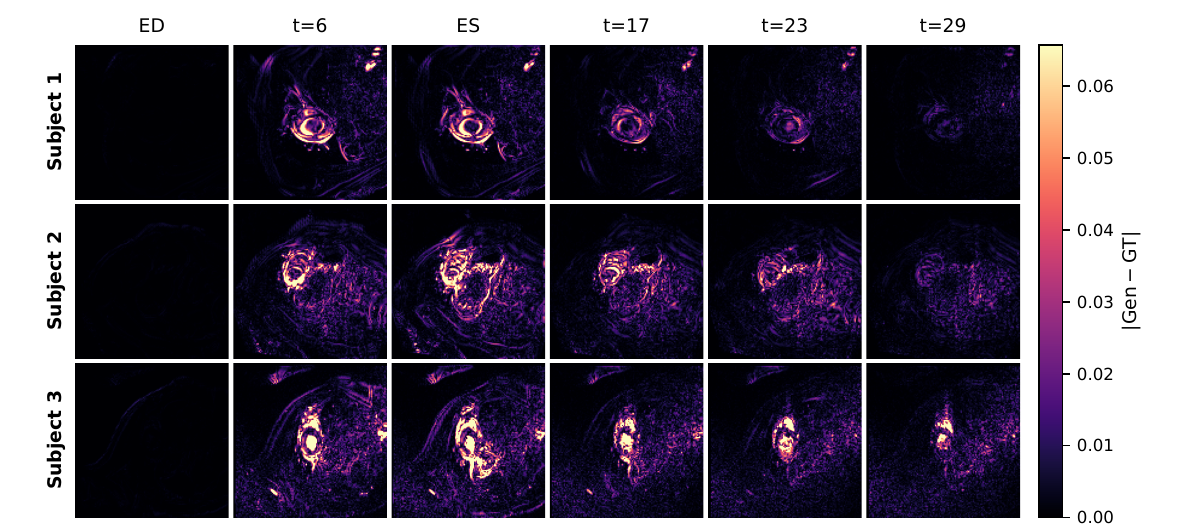}
  \caption{Per-pixel absolute reconstruction error $|\mathrm{Gen}-\mathrm{GT}|$
    for three representative subjects (rows) across six frames spanning the
    cardiac cycle (columns), on the mid-ventricular slice. The error stays near
    zero over the static background and concentrates in a thin band at the
    myocardial boundary, peaking around ES where motion is largest.}
  \label{fig:errormaps}
\end{figure*}

Figure~\ref{fig:qualitative} shows synthesized frames and predicted
displacement fields for representative MINF and NOR patients.
PhaseFlow3D produced sharp, anatomically coherent frames across the full
cardiac cycle, with clearly visible myocardial thickening during systole and
chamber expansion during diastole; the VAE-decoded variant exhibited the
characteristic blur of VAE reconstruction, particularly at fine anatomical
boundaries.
The displacement magnitude maps show smooth, anatomically localized
deformation concentrated around the myocardium, consistent with the spatial
localization enforced by the radial head's dilated mask.
Figure~\ref{fig:errormaps} maps the per-pixel absolute reconstruction error
$|\mathrm{Gen}-\mathrm{GT}|$ for three representative subjects across the cardiac
cycle. The error stays near zero over the static background and concentrates in a
thin band at the myocardial boundary, peaking around ES where motion is largest.

\begin{figure*}[width=\textwidth,pos=!t]
  \centering
  \includegraphics[width=0.95\linewidth]{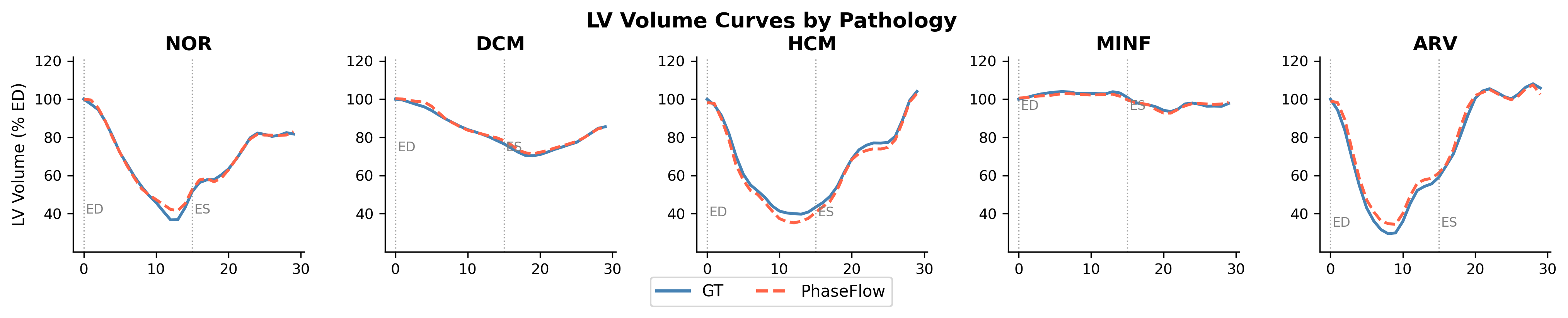}
  \caption{LV volume curves      (\%~of ED volume). Ground-truth (solid)
    vs.\ PhaseFlow3D-predicted (dashed) trajectories for one representative
    patient per ACDC pathology group.}
  \label{fig:volume_curves}
\end{figure*}

Figure~\ref{fig:volume_curves} compares predicted and ground-truth LV volume
curves.
PhaseFlow3D captured the pathology-specific contraction patterns: normal
hearts showed the expected ${\sim}55{-}70\%$ EF, DCM patients exhibited
reduced contraction, and HCM patients showed preserved but hypertrophic
dynamics.
The piecewise phase conditioning correctly reproduced the asymmetric
systole/diastole velocity profile, visible as a steep systolic descent
followed by a shallower diastolic recovery.
Per-pathology qualitative results across all five ACDC groups are provided in
Appendix~\ref{app:qualitative}.

\section{Downstream Applications}
\label{sec:downstream}

The preceding section established that PhaseFlow3D synthesizes sequences that
are faithful in image quality and physiology. This section asks whether the
synthesized sequences and their displacement fields are useful. We evaluate
four applications: substituting synthetic supervision for real supervision in
segmentation and in pathology classification, propagating segmentation labels
through the recovered displacement field, and deriving myocardial strain from
the generated motion. A fifth, auxiliary experiment measures perceptual realism
with a learned discriminator. In all experiments the generator is frozen and
inference is blind: a single real ED volume and the canonical phase template,
with no ES image, ECG, or pathology label. Unless stated otherwise, results are
reported as mean $\pm$ standard deviation over three random seeds, which
reflects seed variability rather than a confidence interval. Protocol details
and full per-fraction tables are given in Appendix~\ref{app:downstream}.

\subsection{Synthetic Supervision for Segmentation}
\label{sec:seg_subst}

We first test whether a synthesized training sample can stand in for a real
one. With the training budget fixed, an increasing fraction of the real
supervision for a 2D segmentation U-Net (LV and myocardium) is replaced by
synthetic supervision, and the network is always evaluated on the real ACDC
test ED and ES frames. Pipeline A performs phase-aligned substitution: a
patient's real ES image and annotation are replaced by the synthesized ES
frame and the ED mask warped to ES. Pipeline B substitutes real ED and ES
slices with synthesized intermediate-phase slices carrying
 pseudo-labels at a fixed total slice budget.

\begin{figure}[pos=t]
  \centering
  \includegraphics[width=0.6\linewidth]{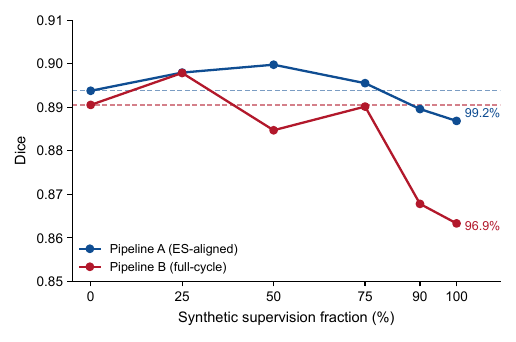}
  \caption{Segmentation substitution. Test Dice (LV and myocardium mean,
    3-seed mean) vs.\ the fraction of real supervision replaced by synthetic
    supervision, for phase-aligned substitution (pipeline A, blue) and
    full-cycle substitution (pipeline B, red). Dashed lines: each pipeline's
    all-real baseline. Right edge: retention, the ratio of fully synthetic to
    all-real Dice.}
  \label{fig:seg_substitution}
\end{figure}

Figure~\ref{fig:seg_substitution} shows the resulting Dice over the
substitution fraction. Under pipeline A, Dice moves from $0.894 \pm 0.005$
with all-real supervision to $0.887 \pm 0.002$ with fully synthetic
supervision, a retention of $99.2\%$, and up to $75\%$ substitution the curve
stays within the seed-noise band of the all-real baseline. Pipeline B retains
$96.9\%$ under full substitution, with a mild decline appearing only beyond
$75\%$. These results indicate that a synthesized ES supervision unit is
nearly interchangeable with a real one.

\subsection{Synthetic Supervision for Pathology Classification}
\label{sec:clf_subst}

The same substitution protocol is applied to five-class pathology
classification on ACDC. A random forest~\cite{breiman2001random} receives five
geometric features in physical units (LV volume, myocardial mass, mean wall
thickness, LV sphericity, and the mass-to-volume ratio) computed from
segmentation masks of the $45$ generator-unseen ACDC patients. Real training samples use the real ED and ES masks;
synthetic training samples use the ED mask warped to intermediate synthesized
phases. Evaluation always uses real features, with patient-stratified
five-fold cross-validation.

\begin{figure}[pos=t]
  \centering
  \includegraphics[width=0.6\linewidth]{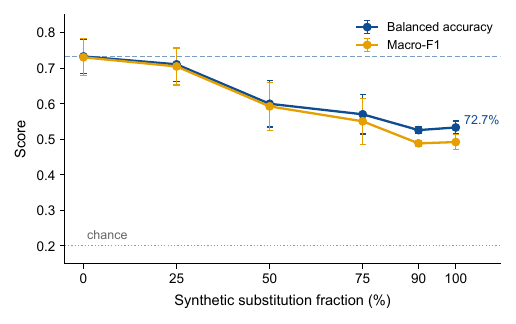}
  \caption{Classification substitution. Balanced accuracy and macro-F1
    (3-seed mean $\pm$ std) vs.\ the fraction of real geometry samples
    replaced by synthetic ones. Dashed line: all-real balanced accuracy;
    dotted line: 5-class chance. Right edge: balanced-accuracy retention.}
  \label{fig:clf_substitution}
\end{figure}

Figure~\ref{fig:clf_substitution} shows the degradation curve. Balanced
accuracy is $0.733 \pm 0.048$ with all-real training and $0.711 \pm 0.048$
when a quarter of the training samples are synthetic, a drop of about two
points. Under full substitution it falls to $0.533 \pm 0.018$, a retention of
$72.7\%$ that remains more than $2.5\times$ the chance level of $0.2$.
Synthetic geometry therefore carries most, though not all, of the
class-discriminative signal, and moderate substitution is nearly free.

\subsection{Label Propagation Through the Displacement Field}
\label{sec:label_prop}

The displacement field itself can transfer annotations. We warp each ACDC test
patient's real ED annotation to ES through the recovered displacement field
and score it against the real ES annotation. Two references bound the task: a
no-motion floor that reuses the ED mask as the ES prediction, and classical
Demons registration~\cite{thirion1998demons}, an oracle upper bound because it
registers the real ED image to the real ES image, which our blind setting
never observes.

\begin{figure}[pos=t]
  \centering
  \includegraphics[width=0.6\linewidth]{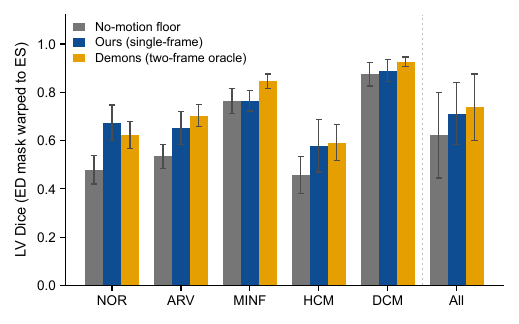}
  \caption{Label propagation. LV Dice of the ED annotation warped to ES, per
    ACDC pathology group and pooled (All), for the no-motion floor, our blind
    single-frame propagation, and the Demons two-frame oracle that sees the
    real ES image. Bars: patient mean; error bars: patient std.}
  \label{fig:label_prop}
\end{figure}

Figure~\ref{fig:label_prop} reports LV Dice on the $30$ test patients. Our
propagation reaches $0.712 \pm 0.129$, recovering most of the interval between
the floor ($0.622 \pm 0.177$) and the two-frame oracle ($0.738 \pm 0.138$)
while using no ES information. In the NOR and ARV groups the blind propagation
even exceeds the oracle. Myocardium Dice is lower ($0.572$ vs.\ $0.700$ for
Demons; Table~\ref{tab:label_prop_struct}), as expected for a thin structure
under an LV-centric prior.

\subsection{Myocardial Strain Analysis}
\label{sec:strain}

Motion analysis is a core promised use of the recovered displacement field. We
compute global circumferential strain (GCS) at peak systole, identified as the
synthesized frame of minimum warped LV volume, from the Green--Lagrange strain
tensor of the displacement field averaged over the warped myocardium
(Appendix~\ref{app:downstream}). The analysis covers all $166$ combined test
patients and remains fully blind, so any group structure must arise from the
single ED volume alone.

\begin{figure}[pos=t]
  \centering
  \includegraphics[width=0.6\linewidth]{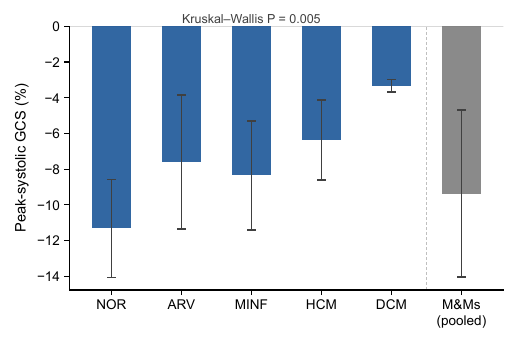}
  \caption{Peak-systolic GCS from the synthesized motion, by ACDC pathology
    group (blue) and for the pooled M\&Ms cohort (grey).
    Bars: group mean; error bars: patient standard deviation. The
    Kruskal--Wallis test is computed across the five ACDC groups.}
  \label{fig:strain_gcs}
\end{figure}

Figure~\ref{fig:strain_gcs} shows the per-patient GCS. The ACDC groups
separate according to expected contractile function: NOR shows the strongest
contraction ($-11.3 \pm 2.7\%$), MINF and ARV are intermediate
($-8.4 \pm 3.0\%$ and $-7.6 \pm 3.7\%$), HCM is reduced ($-6.4 \pm 2.2\%$),
and DCM is weakest ($-3.3 \pm 0.3\%$); a Kruskal--Wallis test across the five
groups gives $H = 14.9$, $p = 0.005$. The pooled M\&Ms cohort
($-9.4 \pm 4.7\%$) falls between NOR and the diseased groups, consistent with
its mixed composition. Strain magnitudes are smaller than clinically reported
values, in line with the residual under-contraction quantified by the EF
analysis in Sec.~\ref{sec:experiments}, so we interpret this experiment as
validating group separation and ordering rather than absolute strain values.

\subsection{Perceptual Realism}
\label{sec:realism}

As an auxiliary check, we train a small convolutional neural network (CNN)
discriminator to separate real cine slices from synthesized ones, where an
area under the receiver operating characteristic (ROC) curve (AUC) near $0.5$
means the two are hard to distinguish.
The discriminator sees intermediate-phase slices of the $45$ generator-unseen
ACDC patients ($1350$ slices per arm) under patient-level five-fold
cross-validation; no tuning is performed in either direction.

\begin{figure*}[width=\textwidth,pos=t]
  \centering
  \includegraphics[width=0.78\linewidth]{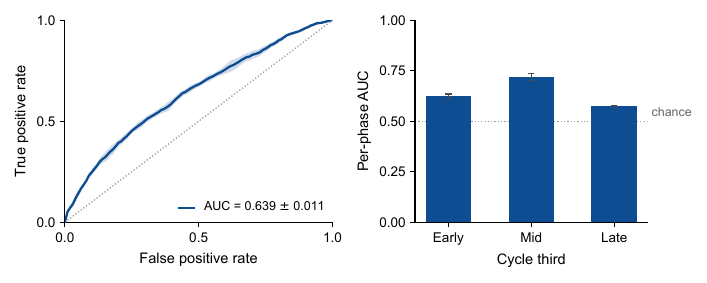}
  \caption{Perceptual realism. Left: pooled out-of-fold ROC of the
    real-vs-synthetic discriminator (3-seed mean, shaded band: seed std);
    the dotted diagonal is chance. Right: AUC by cycle third; mid-systole
    carries most of the residual separability.}
  \label{fig:realism_disc}
\end{figure*}

The discriminator attains an AUC of $0.639 \pm 0.011$
(Figure~\ref{fig:realism_disc}), leaving only mild residual separability above
the chance level of $0.5$.
Per-phase AUC peaks in the mid-systolic third ($0.718$) and falls to $0.574$
late in the cycle, consistent with the mild under-contraction around ES noted
above being the main residual cue. Because real frames pass through the same
temporal resampling as the synthesis grid, any residual preprocessing
difference can only assist the discriminator, so this AUC is a conservative
upper bound on the separability attributable to synthesis.

\section{Discussion}
\label{sec:discussion}

\paragraph{Physiological fidelity and design priors.}
In our comparisons, SimVP and Direct Reg reach higher PSNR yet have larger
EF errors and lower volume-curve $R^2$ than PhaseFlow3D. This is an empirical
finding about physiological accuracy. The perception--distortion
trade-off~\cite{blau2018perception} concerns image distortion and distributional
quality, and does not by itself predict errors in cardiac function.
The metric families answer different questions: PSNR measures how closely a
specific ground-truth sequence is reproduced, FID and FVD measure whether the
generated distribution matches the real one, and EF and volume metrics measure
whether the synthesized heart \emph{beats correctly}.
That PhaseFlow3D more than halves the EF error of the strongest baseline while
conceding only a modest amount of PSNR indicates that task-specific
physiological metrics, rather than pixel accuracy, should drive the evaluation
of cardiac synthesis.
This physiological fidelity follows from our design priors.
The Radial Contraction Decomposition is the component whose removal costs the
most physiological accuracy (A2), whereas the residual is the component whose
removal costs the most image quality (A3), so the decomposition cleanly
separates the two responsibilities.
Beyond accuracy, the radial head offers two practical benefits.
Its global amplitude $A_{\mathrm{pat}}$ is a directly interpretable,
patient-specific readout of contraction magnitude, shaped by the EF loss
during training.
Because the radial direction field is computed analytically from the LV
centroid, the prior also transfers across patients and scanner vendors without
any additional learning.
A second prior, the piecewise cardiac phase, governs how the model is
conditioned at inference.
Its canonical template makes inference fully blind: a single 3D ED volume
suffices, with no ECG, gating, pathology label, or segmentation inference at
deployment.
Ablation A5 confirms that this asymmetric parameterization is what preserves the
volume-curve shape, which a uniform phase fails to recover despite matching
comparable EF accuracy.
The population-level template closely approximates individual ES timing
(Sec.~\ref{sec:phase_validation}), suggesting that the temporal approximation is
useful for the synthesis task on the evaluated cohorts; clinical utility
requires separate validation.

\paragraph{Utility and applications.}
Because inference requires only a single ED volume and no ECG, and because every
synthesized frame is accompanied by an explicit displacement field,
PhaseFlow3D is useful beyond visualization, and
Sec.~\ref{sec:downstream} quantifies this utility directly.
The recovered displacement field supports quantitative cardiac motion analysis
over the full cycle, recovering the expected ordering of contractile function
across pathology groups from a single ED volume, and the synthesis of a
complete sequence from one frame offers a route to recovering missing frames
when only an isolated reference volume survives in retrospective archives.
Synthesized sequences can also substitute for real supervision in downstream
segmentation and disease classification, where paired cine data are scarce,
with near-lossless retention for phase-aligned segmentation supervision and
moderate cost for classification.
Finally, since each synthetic sequence carries a known ground-truth deformation
field, the generated data can serve as a benchmark for validating deformable
registration algorithms and for in-silico trials that require a moving cardiac
model with known motion.

\paragraph{Limitations.}
The EF MAE remains $12.85$ percentage points, indicating substantial residual
error in patient-specific functional prediction. Published inter-modality
agreement limits~\cite{bellenger2000interchangeable} are not directly comparable
with this MAE and do not establish clinical equivalence. Dedicated
reproducibility studies and validation on larger, more diverse cine CMR
cohorts remain necessary.
A single ED volume and a canonical phase also cannot uniquely determine a
patient's true contraction trajectory; generated dynamics remain model-based
predictions rather than observed measurements.
All experiments use ECG-gated cine acquisitions resampled to $D=16$ slices;
very thick or thin slabs may require adapting the depth compression, and
extension to free-breathing acquisitions or to anchor frames other than ED
remains future work.
Finally, pathologies with pronounced non-radial motion, such as severe apical
rotation in HCM, stress the residual pathway and warrant dedicated study.

\section{Conclusion}
\label{sec:conclusion}

In this paper, we developed PhaseFlow3D, a generative framework for synthesizing
4D cine CMR sequences from a single ED volume without any ECG dependency.
The model is generative and can efficiently produce physiologically faithful
cardiac motion for a large number of subjects.
Compared with conventional cine synthesis and registration techniques, the
proposed model generates more physiologically accurate ventricular dynamics
while requiring only a single 3D volume at inference.
In terms of physiological fidelity, it attains the best EF accuracy and the most
faithful LV volume-curve reconstruction among all compared methods.
Besides, the proposed framework synthesises high temporal resolution cardiac
motion sequences together with their underlying displacement fields, which can be
applicable for cardiac motion analysis, for recovering missing frames when only a
single reference volume survives, for data augmentation in downstream
segmentation and disease classification, and for validating deformable
registration algorithms and enabling in-silico trials where a synthetic sequence
with a known deformation field is required.
Our future work will focus on extending this model to free-breathing
acquisitions and to anchor frames other than ED.
Besides, we will validate the framework on larger and more diverse cine CMR
cohorts toward clinical deployment.

\appendix

\section{Excluded Baselines}
\label{app:excluded_baselines}

We detail why three otherwise-natural families of baselines were excluded from
the main comparison.

\paragraph{Registration methods.}
Registration-based methods such as VoxelMorph~\cite{voxelmorph} address a
different task: aligning two \emph{observed} images. In that formulation,
a target frame is required at inference and is unavailable in our
single-ED-volume setting.

\paragraph{ECG-conditioned methods.}
ECG-conditioned approaches~\cite{echopulse, ecgflowcmr} require an
ECG input at inference, which violates our ECG-free setting and would make a
direct comparison unfair in either direction.

\paragraph{EchoLVFM.}
EchoLVFM~\cite{echolv} has a publicly available implementation and uses
MeanFlow-based generation conditioned on observed echocardiographic frames
and a target EF. A direct comparison would require adaptation from 2D
ultrasound to 3D-volume CMR sequences and a harmonized conditioning protocol:
providing a target EF would add information unavailable to PhaseFlow3D at
inference, while removing it would change EchoLVFM's reported setup.

\section{Implementation Details}
\label{app:implementation}

\paragraph{CineVAE3D.}
A true 3D VAE with per-frame encoding: input $(B, 1, D, H, W)$ per frame,
latent $(B, 4, D', h, w) = (B, 4, 4, 16, 16)$.
The encoder downsamples over four levels: the first two halve depth and
in-plane resolution together, whereas the third halves only the in-plane
resolution, with 3D spatial attention applied at the two coarsest levels.
The temporal dimension is handled externally by per-frame encoding; the flow
model performs all temporal reasoning.
Pretrained for 100 epochs ($\mathrm{lr}=10^{-4}$, KL weight $10^{-6}$,
4 frames sampled per volume per step), then frozen.

\paragraph{VelocityUNet4D.}
D$'$-fold trick: input $(B, 4, T, 4, 16, 16) \rightarrow (B, 16, T, 16, 16)$
before each level, unfolded back after.
3D U-Net backbone: base channels 128, multipliers $[1,2]$, two residual blocks
per level.
AdaGN: the flow time $\tau$ is encoded via a sinusoidal embedding and MLP,
added to the temporal-mean phase embedding as the global condition.
Phase cross-attention: $n_{\mathrm{tokens}}=4$, $n_{\mathrm{heads}}=8$,
per frame, at both levels.
The output convolution is zero-initialized.

\paragraph{ImageEncoder3D.}
A lightweight 3D CNN (3 levels, channels $[16, 32, 64]$) processing
$\mathbf{x}_{\mathrm{ED}}$; its multi-scale feature maps are injected as skip
connections into the deformation decoder at matching resolutions.

\paragraph{ImageConditionedDeformDecoder3D.}
3D transposed convolutions (three upsampling steps) from $(4, 4, 16, 16)$ to
$(3, 16, 128, 128)$, with ImageEncoder3D skip features concatenated after each
step and a learnable output scale initialized to $10.0$.

\paragraph{RadialContractionHead.}
GlobalNet: $\mathrm{GAP}(\mathbf{f}) \rightarrow \mathrm{Linear}(64)
\rightarrow \mathrm{SiLU} \rightarrow \mathrm{Linear}(1) \rightarrow \sigma
\rightarrow A_{\mathrm{pat}}$.
LocalNet: a small 3D convolutional upsampler (a $3{\times}3{\times}3$
convolution, two transposed-convolution upsampling stages, and a final
$1{\times}1{\times}1$ convolution with sigmoid), mean-normalized within the
dilated LV ROI so that it only modulates the spatial distribution of
contraction.
TemporalNet: $\mathrm{Linear}(d_{\mathrm{cond}}, 64) \rightarrow \mathrm{SiLU}
\rightarrow \mathrm{Linear}(64, 1)$ per frame, peak-normalized with ED
subtraction.

\paragraph{SpatialTransformer3D.}
Differentiable 3D warping via trilinear grid sampling with border padding; the
ED volume is broadcast to $T$ frames and each per-frame displacement
$\boldsymbol{\varphi}_t \in \mathbb{R}^{3 \times D \times H \times W}$ is
applied independently.

\paragraph{Training details.}
Stage 1: 50 epochs, physiology-only losses.
Stage 2: 100 epochs, all losses, checkpoint selected by validation EF MAE.
Both stages: AdamW ($\beta_1=0.9$, $\beta_2=0.999$, weight decay $10^{-4}$),
cosine annealing from $\mathrm{lr}=10^{-4}$ to $10^{-6}$, batch size 1, AMP
mixed precision.
$\alpha=0.1$ for $\mathbf{z}_0$ construction; $K=7$ scaling-and-squaring
steps.

\section{Data Preprocessing}
\label{app:preprocessing}

We describe the full preprocessing pipeline applied to the ACDC and M\&Ms
cine CMR studies. One offline stage fixes the patient-level splits; a
per-sample transform is then applied on the fly inside the \texttt{Dataset}.
Algorithm~\ref{alg:preprocess} gives the exact sequence of operations.

\begin{algorithm}[h]
\caption{Preprocessing (per patient; 3D volume sequence)}
\label{alg:preprocess}
\begin{algorithmic}[1]
\STATE \textbf{Offline (once):} ACDC: collect all $150$ patients, read
  pathology \texttt{Group} from each \texttt{Info.cfg}, stratified split by
  group into train/val/test $=105/15/30$ with seed $42$
  $\to$ \texttt{split.json}.
  M\&Ms: retain the challenge split assignments for the included cases
  (Labeled Training/Validation/Testing $=150/34/136$).
\STATE \textbf{Per sample} (one patient):
\STATE \quad Load 4D NIfTI $(H, W, Z, T_{\mathrm{orig}})$; read ED/ES frame
  indices (ACDC: \texttt{Info.cfg}; M\&Ms: challenge metadata).
\STATE \quad M\&Ms only: circularly shift the time axis so that ED becomes
  frame $0$ (ACDC sequences already start at ED); remap segmentation labels
  to the ACDC convention.
\STATE \quad Reorder to $(T_{\mathrm{orig}}, 1, Z, H, W)$.
\STATE \quad Spatial: bilinear resize each slice to $128 \times 128$.
\STATE \quad Depth: center pad (zeros) or center crop $Z \to D = 16$.
\STATE \quad Temporal: linear resample $T_{\mathrm{orig}} \to T = 30$ frames;
  rescale the ES index accordingly.
\STATE \quad Intensity: per-patient min--max normalize the resampled
  sequence to $[-1, 1]$ (a single $[v_{\min}, v_{\max}]$ range shared by all
  frames of the sequence).
\STATE \quad Phase: piecewise linear
  $\phi$ (ED${}=0 \to$ ES${}=\pi \to$ last frame${}=2\pi$) from the ED/ES
  indices, resampled to $T$; encode as $[\sin\phi_t, \cos\phi_t]$.
\STATE \quad Masks: binary LV cavity masks at ED and ES (label $3$),
  nearest-neighbor resized and depth-padded/cropped identically to the image
  volume.
\STATE \textbf{return} $(\text{cine}, \mathbf{x}_{\mathrm{ED}},
  \phi_{\mathrm{seq}}, \text{es\_idx}, \mathbf{m}_{\mathrm{ED}},
  \mathbf{m}_{\mathrm{ES}})$.
\end{algorithmic}
\end{algorithm}

No intensity augmentation, cropping, or registration is applied; the only
spatial operations are an isotropic in-plane resize and a center depth
pad/crop. Intensity normalization is performed per patient so that all frames
share one intensity range, preserving relative temporal contrast. The
stratified ACDC split and the M\&Ms subset partitions are reused unchanged
across all models (main and baselines) to ensure an identical evaluation
protocol.

\section{Evaluation Metrics}
\label{app:metrics}

We formally define the metrics used in the main paper and state the
motivation for each. Let $\hat{\mathbf{x}}_t$ and $\mathbf{x}_t$ denote the
generated and ground-truth frame at time $t$ ($t=1,\dots,T$), with voxel
intensities rescaled to $[0,1]$.

\paragraph{Image and video quality.}
\begin{itemize}\setlength{\itemsep}{.2em}
  \item \textbf{PSNR} (dB), averaged over frames:
    $\mathrm{PSNR}=10\log_{10}\!\big(1/\mathrm{MSE}(\hat{\mathbf{x}}_t,\mathbf{x}_t)\big)$.
    Motivation: standard pixel-level fidelity measure; penalizes reconstruction
    error directly.
  \item \textbf{SSIM}~\cite{wang2004ssim}:
    $\mathrm{SSIM}=\frac{(2\mu_{\hat{x}}\mu_x+c_1)(2\sigma_{\hat{x}x}+c_2)}
    {(\mu_{\hat{x}}^2+\mu_x^2+c_1)(\sigma_{\hat{x}}^2+\sigma_x^2+c_2)}$.
    Motivation: captures structural/luminance/contrast agreement, correlating
    better with perceived quality than PSNR for anatomical structure.
  \item \textbf{LPIPS}~\cite{zhang2018lpips}: distance between deep features of
    $\hat{\mathbf{x}}_t$ and $\mathbf{x}_t$.
    Motivation: perceptual similarity that is sensitive to texture realism,
    which pixel metrics miss.
  \item \textbf{FID}~\cite{heusel2017fid}:
    $\lVert\mu_r-\mu_g\rVert^2+\mathrm{Tr}\!\big(\Sigma_r+\Sigma_g
    -2(\Sigma_r\Sigma_g)^{1/2}\big)$ over Inception features of real ($r$) and
    generated ($g$) frames.
    Motivation: measures distributional realism of the generated frames as a
    set, complementing per-frame metrics.
  \item \textbf{FVD}~\cite{unterthiner2019fvd}: the same Fr\'echet distance
    computed over I3D video features of complete sequences.
    Motivation: sensitive to temporal coherence and motion statistics that
    frame-level metrics cannot capture.
\end{itemize}

\paragraph{Physiological fidelity.}
Let $V_t$ and $\hat{V}_t$ be the ground-truth and predicted LV cavity volume
at frame $t$.
Under the primary protocol reported in Table~\ref{tab:main}, $\hat{V}_t$ for
PhaseFlow3D is obtained by warping the ED segmentation mask through the
predicted displacement field, so the segmentation network is never applied to
generated frames; for every baseline, $\hat{V}_t$ is obtained by segmenting
the generated frames with the same frozen SegUNet.
\begin{itemize}\setlength{\itemsep}{.2em}
  \item \textbf{EF~MAE} (\%): absolute error of ejection fraction
    $\mathrm{EF}=(V_{\max}-V_{\min})/V_{\max}$.
    Motivation: EF is the primary clinical index of cardiac function; a direct
    measure of physiological correctness.
  \item \textbf{Vol~Corr}: Pearson correlation between $\{\hat{V}_t\}$ and
    $\{V_t\}$.
    Motivation: tests whether the \emph{shape} of the volume--time trajectory
    is reproduced, independent of scale.
  \item \textbf{Vol~$R^2$}: coefficient of determination
    $1-\sum_t(\hat{V}_t-V_t)^2/\sum_t(V_t-\bar{V})^2$.
    Motivation: penalizes both bias and variance of the volume curve; negative
    values expose predictions worse than the mean.
  \item \textbf{Vol~MAE}: mean absolute per-frame volume error, normalized by
    ED volume.
    Motivation: a scale-aware complement to the correlation-based measures.
\end{itemize}

\paragraph{Cross-protocol validation of EF.}
The primary protocol measures PhaseFlow3D and the baselines through different
pathways, whose error sources differ.
To verify that this asymmetry does not drive the ranking in
Table~\ref{tab:main}, we recomputed the EF MAE on the same $166$ test patients
under two unified protocols (Table~\ref{tab:ef_protocols}).
Under the image-based protocol, the generated frames of every method,
including ours, are segmented by the same frozen SegUNet, and the reference EF
is derived from the expert annotations.
Both unified protocols take the annotations as the reference, which keeps the
reference independent of the segmentation network that enters the predicted
volumes.
In Table~\ref{tab:main}, the baseline EF is instead referenced against the
SegUNet volume curve measured on the real frames, because the per-frame
volume metrics require a reference at every frame whereas the annotations
exist only at ED and ES; this accounts for the small offsets between the two
tables.
Under the mask-propagation protocol, the ED annotation is warped through each
method's own displacement field, following exactly the transformation that
produced its frames.
The latter protocol is defined only for methods that expose a displacement
field in image space; among the baselines, only DragNet does, through its
autoregressive spatial-transformer warps.
Direct Reg regresses voxel intensities and exposes no displacement field.
LFDM displaces only its latent grid and reaches image space through a learned
decoder, so no pixel-space field exists to propagate.
The remaining baselines synthesize intensities without an explicit motion
representation.

\begin{table}[pos=h]
  \centering
  \caption{EF MAE (\%, patient mean $\pm$ std, $n{=}166$) under the two
    unified measurement protocols, both referenced against the expert
    annotations. ``---'': the method exposes no pixel-space displacement
    field, so mask propagation is undefined. ConvLSTM is omitted because its
    EF is undefined under either protocol (cf.\ Table~\ref{tab:main}). For
    GAN-cMRI, $38$ cases whose generated frames yield an empty LV cavity are
    excluded ($n{=}128$).}
  \label{tab:ef_protocols}
  \small
  \begin{tabular}{lcc}
    \toprule
    Method & Image-based & Mask propagation \\
    \midrule
    GAN-cMRI~\cite{gancmri}        & $38.68 \pm 18.07$ & --- \\
    LFDM~\cite{lfdm}               & $29.67 \pm 15.76$ & --- \\
    CVAE                           & $31.70 \pm 18.55$ & --- \\
    DragNet~\cite{dragnet}         & $36.49 \pm 16.43$ & $37.73 \pm 16.80$ \\
    SimVP~\cite{simvp}             & $26.22 \pm 18.08$ & --- \\
    Direct Reg                     & $29.71 \pm 17.66$ & --- \\
    \midrule
    \textbf{PhaseFlow3D (Ours)}    & $\mathbf{14.55 \pm 12.53}$ & $\mathbf{12.85 \pm 11.03}$ \\
    \bottomrule
  \end{tabular}
\end{table}

Three observations follow.
First, under the unified image-based protocol PhaseFlow3D attains an EF MAE of
$14.55\%$ against $26.22\%$ for the best baseline, so the advantage in
Table~\ref{tab:main} is not an artefact of the measurement pathway.
Second, for DragNet, the one baseline that admits both protocols, the two
pathways closely agree ($37.73\%$ versus $36.49\%$), confirming that its EF
error reflects under-estimated motion rather than segmentation error on
generated frames.
Third, applied to real frames and compared against the expert annotations,
the segmentation network alone incurs an EF MAE of $5.48\% \pm 5.33\%$, which
bounds the measurement noise shared by all image-based entries and lies well
below every inter-method gap.
For completeness, referencing the segmenter's volumes on real frames instead
of the annotations changes the image-based EF MAE of PhaseFlow3D from
$14.55\%$ to $14.34\%$.

\section{Metric Variability}
\label{app:statistics}

All reported numbers are obtained from a single training run under fixed seed
$42$; to characterize the distribution of performance beyond point estimates,
Table~\ref{tab:variability} reports both the per-patient mean $\pm$ standard
deviation and the median with interquartile range (IQR) of PhaseFlow3D over
the joint ACDC$+$M\&Ms test set ($n{=}166$ patients).
Image-quality metrics are tightly concentrated (e.g.\ SSIM $0.927\pm0.012$).
The physiological metrics are more dispersed and heavy-tailed: because
Vol~$R^2$ is unbounded below, a small number of outlier patients pull its mean
toward zero and inflate its standard deviation, so the median and IQR
(Vol~$R^2$ median $0.567$, IQR $[-0.058,\,0.797]$) give a more representative
summary.
This spread reflects genuine between-patient variation in cardiac motion and
pathology severity rather than estimator noise.

\begin{table}[pos=h]
  \centering
  \caption{Per-patient dispersion of PhaseFlow3D on the joint ACDC$+$M\&Ms
    test set ($n{=}166$ patients), reported as mean $\pm$ standard deviation
    and as median with interquartile range.}
  \label{tab:variability}
  \small
  \begin{tabular}{lcc}
    \toprule
    Metric & Mean $\pm$ Std & Median [IQR] \\
    \midrule
    \multicolumn{3}{l}{\emph{Image quality}} \\
    \quad PSNR (dB)        & $34.97 \pm 3.24$  & $34.67$ $[32.66,\,37.32]$ \\
    \quad SSIM             & $0.927 \pm 0.012$ & $0.926$ $[0.917,\,0.936]$ \\
    \quad LPIPS            & $0.017 \pm 0.006$ & $0.015$ $[0.013,\,0.021]$ \\
    \midrule
    \multicolumn{3}{l}{\emph{Physiological fidelity}} \\
    \quad EF MAE (\%)      & $12.85 \pm 11.03$ & $9.49$ $[4.58,\,18.41]$ \\
    \quad Vol Corr         & $0.845 \pm 0.166$ & $0.910$ $[0.810,\,0.952]$ \\
    \quad Vol $R^2$        & $0.054 \pm 1.510$ & $0.567$ $[-0.058,\,0.797]$ \\
    \quad Vol MAE          & $0.110 \pm 0.060$ & $0.094$ $[0.063,\,0.142]$ \\
    \bottomrule
  \end{tabular}
\end{table}

\section{Latent Scale and Anchor Perturbation}
\label{app:latent_scale}

The anchor noise scale $\alpha$ in the anchor latent construction
$\mathbf{z}_0 = \mathrm{repeat}_T\!\big(E_{\mathrm{vae}}(\mathbf{x}_{\mathrm{ED}})\big)
+ \alpha\boldsymbol{\varepsilon}$
(with $\boldsymbol{\varepsilon} \sim \mathcal{N}(\mathbf{0},\,\mathbf{I})$) is
only meaningful relative to the scale of the latent space itself. We therefore
measure the empirical statistics of the frozen CineVAE3D latents over the
ACDC$+$M\&Ms training and validation splits (Table~\ref{tab:latent_scale}).
Latents are taken as the posterior mode $\boldsymbol{\mu}$, matching the
training configuration; the full sequence latent has dimension
$D_{\mathrm{lat}} = C \times T \times D' \times h \times w
= 4 \times 30 \times 4 \times 16 \times 16 = 122{,}880$.

\begin{table}[pos=h]
  \centering
  \caption{Empirical latent statistics ($\sigma$ = per-element standard
    deviation, averaged over patients). The two splits agree closely,
    indicating the latent scale is a stable property of the frozen encoder
    rather than a split artifact.}
  \label{tab:latent_scale}
  \small
  \begin{tabular}{lcc}
    \toprule
    Quantity & Train ($n{=}255$) & Val ($n{=}49$) \\
    \midrule
    $\sigma$ of $\mathbf{z}_1$                                    & 0.800 & 0.794 \\
    $\sigma$ of anchor $E_{\mathrm{vae}}(\mathbf{x}_{\mathrm{ED}})$ & 0.798 & 0.793 \\
    $\sigma$ of $\mathbf{z}_1 - \mathrm{repeat}_T(\cdot)$          & 0.131 & 0.122 \\
    \midrule
    $\alpha / \sigma(\mathbf{z}_1)$                                & 12.5\% & 12.6\% \\
    $\alpha / \sigma(\mathbf{z}_1 - \mathrm{repeat}_T(\cdot))$     & 76.6\% & 82.1\% \\
    \bottomrule
  \end{tabular}
\end{table}

\paragraph{Geometry of the perturbation.}
At the vector level the anchor noise is not small relative to the motion
signal.
On the validation split the transport distance is
$\lVert \mathbf{z}_1 - \mathbf{z}_0 \rVert = 42.70 \pm 9.70$, the noise norm
is $\lVert \alpha\boldsymbol{\varepsilon} \rVert = 35.04 \pm 0.07$ (matching
the theoretical $\alpha\sqrt{D_{\mathrm{lat}}} = 35.05$), and the perturbed
and clean transport directions differ by $40.2^\circ \pm 6.2^\circ$.
Two observations explain why a perturbation of this magnitude is
nevertheless benign.
First, the component of the noise along the motion direction is negligible.
Its projection onto $\mathbf{z}_1 - \mathbf{z}_0$ is $0.073 \pm 0.056$,
i.e.\ $0.18\%$ of the transport distance, matching the prediction that for
isotropic $\boldsymbol{\varepsilon}$ this projection follows
$\mathcal{N}(0, \alpha^2)$ independently of $D_{\mathrm{lat}}$
($\mathbb{E}|{\cdot}| = \alpha\sqrt{2/\pi} = 0.080$).
The noise therefore introduces no appreciable bias in the transport length
along the motion direction.
Second, the large orthogonal component changes the start point but not the
supervision target.
The perturbed anchor $\mathbf{z}_0$ is itself the input to the velocity
network, and the endpoint $\mathbf{z}_1$ is held fixed, so each noise draw
defines a valid training pair from a different start point to the same
target.
The noise thus acts as an augmentation of the anchor rather than as
unobservable corruption: the network is trained to transport the entire
cone of perturbed anchors to the same $\mathbf{z}_1$, and at inference it
observes the noisy anchor it must compensate for.
On this basis $\alpha = 0.1$ is a safe choice.
The training split gives consistent values ($45.79 \pm 13.50$,
$35.06 \pm 0.07$, projection $0.19\%$, angle $38.8^\circ \pm 7.9^\circ$).

\section{Extended Qualitative Results}
\label{app:qualitative}

\begin{figure}[pos=t]
  \centering
  \includegraphics[width=\linewidth,height=0.88\textheight,keepaspectratio]{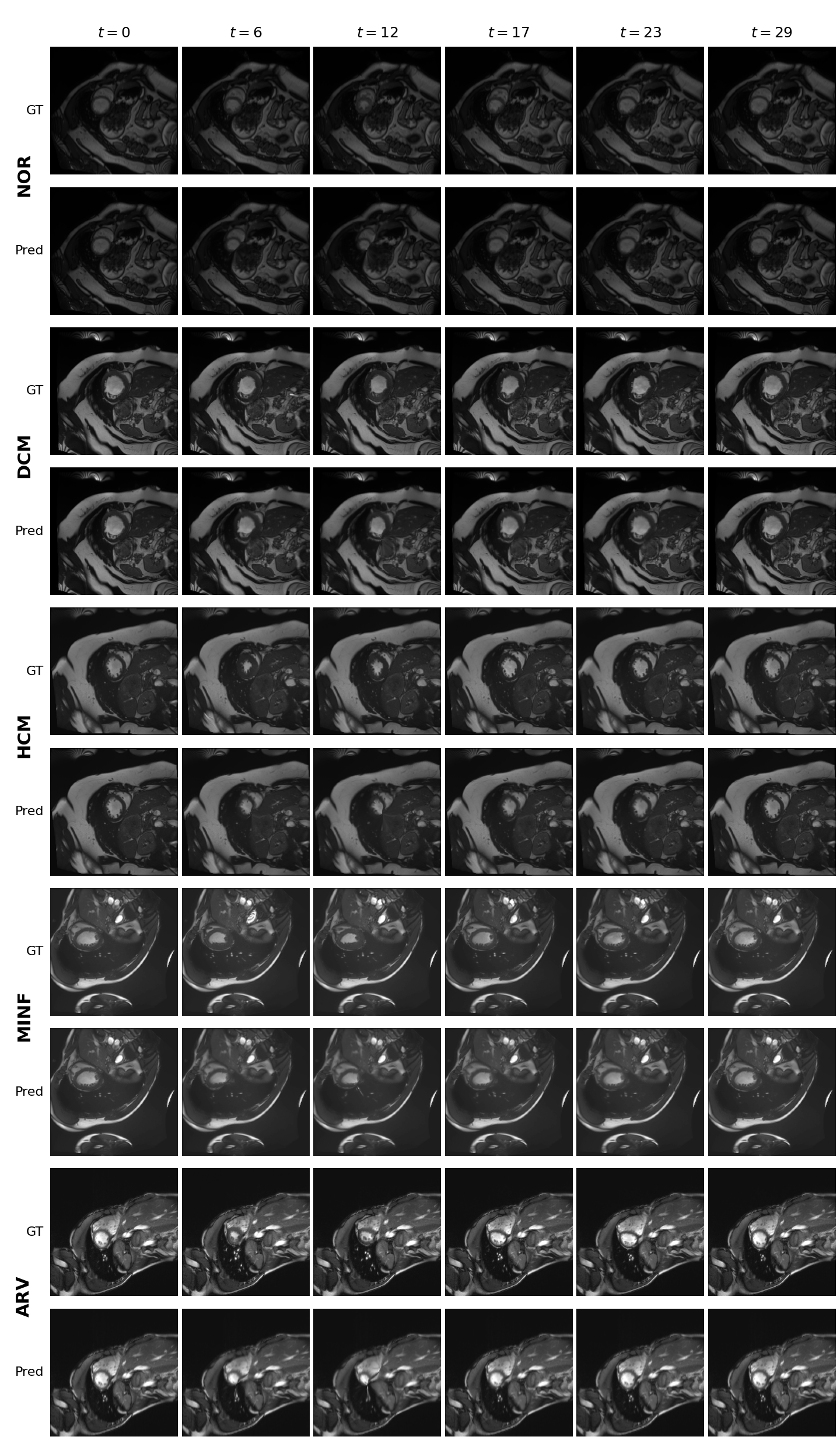}
  \caption{Per-pathology qualitative results (one representative patient per
    group). \emph{Top row}: ground truth; \emph{bottom row}: PhaseFlow3D
    synthesis. ED frame ($t=0$, leftmost) is the sole model input.}
  \label{fig:qualitative_grid}
\end{figure}

Figure~\ref{fig:qualitative_grid} presents representative generated cine
sequences for one patient from each ACDC pathology group (mid-ventricular
slice).
Each group occupies two rows: the top row shows the ground-truth sequence and
the bottom row shows the PhaseFlow3D synthesis, at six evenly-spaced frames
spanning the full cardiac cycle.
The leftmost frame (ED, $t=0$) is the only input to the model at inference
time; all subsequent frames are synthesized from the canonical phase template
and the predicted 3D displacement field.

PhaseFlow3D faithfully captures the characteristic motion signatures of each
pathology: normal wall-thickening in NOR, large-cavity contraction in DCM,
asymmetric septal hypertrophy in HCM, regional wall-motion abnormality in MINF,
and preserved right ventricular dynamics in ARV.

\section{Downstream Experiment Protocols and Full Results}
\label{app:downstream}

This appendix records the protocols and complete per-fraction results of the
downstream experiments in Sec.~\ref{sec:downstream}. All experiments use the
frozen main checkpoint under blind canonical-phase inference. Substitution
experiments use seeds $\{42, 43, 44\}$ and report the seed mean $\pm$ standard
deviation.

\paragraph{Segmentation substitution (Sec.~\ref{sec:seg_subst}).}
The segmentation network is a 2D U-Net trained for $40$ epochs per
configuration on ACDC training patients, with classes LV and myocardium and
the RV mapped to an ignore label in synthetic pseudo-labels. In pipeline A the
substitution unit is a patient's ES supervision pair (image and annotation),
replaced by the synthesized ES frame and the real ED annotation warped to ES.
In pipeline B the real pool holds ED and ES slices while the synthetic pool
holds eight intermediate synthesized phases per patient with warped
pseudo-labels, and substitution proceeds at a fixed total slice budget.
Evaluation uses the real ACDC test ED and ES frames in every configuration.
Table~\ref{tab:downstream_b3} lists the full curves.

\begin{table}[pos=h]
  \centering
  \caption{Segmentation substitution: test Dice (LV and myocardium mean,
    3-seed mean $\pm$ std) at each synthetic supervision fraction.}
  \label{tab:downstream_b3}
  \small
  \begin{tabular}{lcc}
    \toprule
    Synthetic fraction & Pipeline A & Pipeline B \\
    \midrule
    $0\%$   & $0.894 \pm 0.005$ & $0.891 \pm 0.008$ \\
    $25\%$  & $0.898 \pm 0.003$ & $0.898 \pm 0.005$ \\
    $50\%$  & $0.900 \pm 0.001$ & $0.885 \pm 0.003$ \\
    $75\%$  & $0.896 \pm 0.000$ & $0.890 \pm 0.006$ \\
    $90\%$  & $0.890 \pm 0.003$ & $0.868 \pm 0.011$ \\
    $100\%$ & $0.887 \pm 0.002$ & $0.863 \pm 0.009$ \\
    \midrule
    Retention & $99.2\%$ & $96.9\%$ \\
    \bottomrule
  \end{tabular}
\end{table}

\paragraph{Classification substitution (Sec.~\ref{sec:clf_subst}).}
The classifier is a random forest with $400$ trees under patient-stratified
five-fold cross-validation; a patient's samples never cross folds, and test
folds always use features from the real ED and ES annotations, aggregated per
patient by the mean predicted probability. Real training samples contribute
two feature vectors per patient (ED and ES); synthetic samples contribute
eight, one per intermediate synthesized phase, computed from the warped ED
annotation. All five features are computed in physical units from the voxel
spacing. Table~\ref{tab:downstream_c2} lists the full curves.

\begin{table}[pos=h]
  \centering
  \caption{Classification substitution: five-class balanced accuracy and
    macro-F1 (3-seed mean $\pm$ std) at each synthetic substitution fraction.}
  \label{tab:downstream_c2}
  \small
  \begin{tabular}{lcc}
    \toprule
    Synthetic fraction & Balanced accuracy & Macro-F1 \\
    \midrule
    $0\%$   & $0.733 \pm 0.048$ & $0.731 \pm 0.052$ \\
    $25\%$  & $0.711 \pm 0.048$ & $0.705 \pm 0.052$ \\
    $50\%$  & $0.600 \pm 0.065$ & $0.593 \pm 0.067$ \\
    $75\%$  & $0.570 \pm 0.055$ & $0.551 \pm 0.064$ \\
    $90\%$  & $0.526 \pm 0.010$ & $0.489 \pm 0.009$ \\
    $100\%$ & $0.533 \pm 0.018$ & $0.492 \pm 0.021$ \\
    \bottomrule
  \end{tabular}
\end{table}

\paragraph{Label propagation (Sec.~\ref{sec:label_prop}).}
The experiment is deterministic (no training, no seeds). Our arm warps the
multiclass ED annotation through the recovered displacement field to the
canonical ES frame with nearest-neighbour interpolation. The Demons oracle
runs fast symmetric-forces Demons registration~\cite{thirion1998demons} for
$50$ iterations from the real ED image to the real ES image after histogram
matching, then applies the resulting field to the ED annotation.
Table~\ref{tab:label_prop_struct} reports both structures.

\begin{table}[pos=h]
  \centering
  \caption{Label propagation on the ACDC test set ($n{=}30$): Dice of the ED
    annotation warped to ES (patient mean $\pm$ std) per structure and arm.}
  \label{tab:label_prop_struct}
  \small
  \begin{tabular}{lccc}
    \toprule
    Structure & No-motion floor & Ours (single-frame) & Demons (oracle) \\
    \midrule
    LV         & $0.622 \pm 0.177$ & $0.712 \pm 0.129$ & $0.738 \pm 0.138$ \\
    Myocardium & $0.486 \pm 0.164$ & $0.572 \pm 0.121$ & $0.700 \pm 0.093$ \\
    \bottomrule
  \end{tabular}
\end{table}

\paragraph{Strain analysis (Sec.~\ref{sec:strain}).}
For each patient, the peak-systolic frame $t^{\ast}$ is the synthesized frame
of minimum warped LV volume. Because the network predicts the backward
sampling map (Eq.~\eqref{eq:warp}), the identity plus the spatial gradient of
the displacement at $t^{\ast}$ is the inverse deformation gradient, evaluated
at frame-$t^{\ast}$ coordinates. This matrix is inverted at each voxel to
obtain the forward deformation gradient, from which the Green--Lagrange
strain tensor is computed. The tensor is projected per slice onto the
circumferential direction defined relative to the LV centroid, and GCS is the
robust mean over the warped myocardium with strain magnitudes above $1$
excluded as numerical outliers. Radial and longitudinal projections are
computed but not reported, for the reasons given in Sec.~\ref{sec:strain}.

\paragraph{Realism discriminator (Sec.~\ref{sec:realism}).}
The discriminator is a four-block strided CNN trained for eight epochs with
learning rate $10^{-3}$ and batch size $64$ on $128 \times 128$ slices. Per
patient and arm, six intermediate-phase frames (excluding ED) and five
LV-containing slices enter the pools, giving $1350$ slices per arm over the
$45$ patients. Folds are split at the patient level, so a patient's real and
synthetic slices share a fold, and the reported AUC is computed on pooled
out-of-fold predictions.

\FloatBarrier

\printcredits

\bibliographystyle{model1-num-names}
\bibliography{refs}

\end{document}